%% file: main.tex
\documentclass[sigconf]{acmart} 
\renewcommand\footnotetextcopyrightpermission[1]{} 
\usepackage{subcaption}
\usepackage{graphics}
\usepackage{amsmath}
\usepackage{array}
\usepackage{multirow}
\usepackage{algorithm, algorithmic}%
\usepackage[T1]{fontenc}
\usepackage[textsize=tiny]{todonotes} 

\newcommand{\specialcell}[2][c]{%
	\begin{tabular}[#1]{@{}c@{}}#2\end{tabular}}

\AtBeginDocument{%
  \providecommand\BibTeX{{%
    \normalfont B\kern-0.5em{\scshape i\kern-0.25em b}\kern-0.8em\TeX}}}

\let\originalparagraph\paragraph
\renewcommand{\paragraph}[1]{\originalparagraph{\textbf{#1}}}

\usepackage{xcolor}
\begin{document}

\title{An Extensive Study of User Identification via Eye Movements across Multiple Datasets}

%

\author{Sahar Mahdie Klim Al Zaidawi}
\email{saharmah@cs.uni-bremen.de}
\affiliation{%
	\institution{Database Lab,\\ University of Bremen\,, Germany}
}

\author{Martin H.U. Prinzler}
\email{martin.prinzler@cs.uni-bremen.de}
\affiliation{%
	\institution{Database Lab,\\ University of Bremen\,, Germany}
}

\author{Jonas L{\"u}hrs}
\email{luehrsj@uni-bremen.de}
\affiliation{%
	\institution{Database Lab,\\ University of Bremen\,, Germany}
}

\author{Sebastian Maneth}
\email{maneth@cs.uni-bremen.de}
\affiliation{%
	\institution{Database Lab,\\ University of Bremen\,, Germany}
}

\renewcommand{\shortauthors}{}
\input{content/abstract}

\maketitle
\pagestyle{plain} 
\input{content/introduction}

\input{content/Proposed_System.tex}

\input{content/User_Prediction_Experiments}

\input{content/Discussion}

\input{content/Conclusion}

\input{content/Acknowledgment}

\bibliographystyle{ACM-Reference-Format}
\bibliography{userref}


\end{document}

%% file: content/abstract.tex
\begin{abstract}
Several studies have reported that biometric identification based on eye movement characteristics can be used for authentication.
This paper provides an extensive study of user identification via eye movements across multiple datasets based on an improved version of method originally proposed by George and Routray. 
We analyzed our method with respect to several factors that affect the identification accuracy, such as the type of stimulus, the IVT parameters (used for segmenting the trajectories into fixation and saccades), adding new features such as higher-order derivatives of eye movements, the inclusion of blink information, template aging, age and gender.
We find that three methods namely selecting optimal IVT parameters, adding higher-order derivatives features and including an additional blink classifier have a positive impact on the identification accuracy.
The improvements range from a few percentage points, up to an impressive 9\,\% increase on one of the datasets.
\end{abstract}

\keywords{Eye Tracking;
	Eye Movement Biometrics; 
	User Identification;
	Machine Learning;
	IVT algorithm
}

%% file: content/introduction.tex
\section{Introduction}
\label{sec:introduction}

Eye movements can comprise rich and sensitive information about an individual, 
including biometric identity, gender, age, ethnicity, personality traits, 
drug consumption habits, moods and emotions, skills, preferences, 
cognitive processes, and physical and mental health conditions~\cite{kroger2019does}.
Eye movements are an excellent predictor of human desires and focus.
They are inextricably linked with human cognitive and perceptual processes.

Eye tracking is a method to record a persons's eye movements.
Today, a vast number of different eye tracking devices are available and
have been used by researchers to produce many high-quality datasets.
These datasets are analyzed in different contexts; we only mention a few examples:
disorder detection~\cite{armstrong2012eye, benfatto2016screening, billeci2017integrated}, 
gaming~\cite{lin2004design, alkan2007studying, lankes2020gazing}, 
gender prediction~\cite{sargezeh2019gender, moss2012eye, DBLP:conf/icmi/ZaidawiPSZM20}, and 
user identification~\cite{DBLP:conf/eccv/KasprowskiO04, george2016score, DBLP:journals/ivc/RigasK17, DBLP:conf/chi/SchroderZPMZ20}

Let us consider the state-of-the-art approaches to user identification via eye movement data.
In the 2015 BioEye competition~\cite{DBLP:journals/ivc/RigasK17} 
different systems for user identification were competing.
The winning system is based on first segmenting the eye tracking trajectory into fixations and saccades
using the ``Velocity-Threshold Identification'' (IVT) method (see e.g.~\cite{DBLP:conf/etra/SalvucciG00}, where this methods is analyzed and some prior references are given).
Their version of IVT uses two parameters which are set to certain default values.
It was not investigated, whether other settings of these parameters would give higher identification accuracies.
Most of the previous studies of eye movement biometrics have tested their approach using a few number of participants, utilizing either one or two datasets e.g.~\cite{george2016score,jager2019deep, krishna2019multimodal}.
Furthermore, since most prior studies train and assess their models on datasets compiled over a short time span, the permanence of eye movements remains unexamined.
Our study fills these gaps by assessing these methods with datasets containing a larger number of participants 
and some that were not utilized for eye movement biometrics.

We use two datasets from the 2015 Bioeye competition~\cite{DBLP:journals/ivc/RigasK17}.
They consists of 153~participants which looked at two different stimuli: random moving dots (RAN) and a poem (TEX).
The third biometrics dataset is about a visual searching task (VST)~\cite{li2018biometrictask} and comprises 58~participants. 
The final dataset is known as the gaze on faces (GOF) and contains 378~participants.
The GOF dataset was used to identify ``scanning strategies'' that are different for men than for women~\cite{coutrot2016face}.
The entire data utilized in our study provides a high number of participants and a broad range of age groups and stimuli.

The main contribution of this paper is an extensive study of user identification via eye movements across four different datasets.
We investigate the effect of stimuli, gender, and age on user identification.
Additionally, we propose the following improvements:
\begin{itemize} 
	\item {We optimize the IVT parameters which can increase the accuracy 
by~3\,\% for RAN, by 2\,\% for TEX, by 5\,\% for GOF, and by 9\,\% for VST datasets}.
	\item {Adding higher derivative features can increase the accuracy by 2\,\% for RAN, by 1\,\% for TEX and VST, and by 3\,\% for GOF datasets}.
	\item {Adding blinking features can increase the accuracy by 1\,\% for RAN and VST, by 0.5\,\% for TEX datasets}.
	\item {Combining the above three methods of improvement (IVT parameters, higher order derivative features, blink classifier) can increase the accuracy by 4\,\% for RAN, by 3\,\% for TEX, and by 9\,\% for VST datasets}.
\end{itemize}	

\textbf{Related Work.}\quad
%
There is vast and fast growing literature on eye tracking biometrics.
Some surveys that provide a good overview are~\cite{Esf16,kroger2019does,DBLP:journals/ivc/RigasK17,DBLP:journals/prl/GaldiNRW16}.
When comparing results, it is important to keep in mind that the fewer participants there are,
the higher the prediction accuracies will be; thus, it is hard to compare results that have different
numbers of participants.
The winner of the 2015 BioEye competition~\cite{george2016score} achieves an accuracy of~89\,\% (over a single run) over 153 participants by computing fixation and saccade based features and using them with an RBFN based classifier;
this was improved in~\cite{DBLP:conf/chi/SchroderZPMZ20} to an accuracy of 94.1\,\% from one run (92.6\,\% over 50 runs 
as shown in the current paper), 
basically by adding more features. 
A different biometrics approach has been implemented in~\cite{li2018biometrictask} for 
the VST dataset over 58 participants.
They split their stimuli (images) into different grids and consider the
frequencies of gaze points per grid component
(using the Gabor wavelet transform). 
The highest accuracy in this work is 97.35\,\% (termed as rank-1 identification rate) using 10~runs
(data was split into 70\,\% train and 30\,\% test) with their Fixation Density Map (FDM) method.

%% file: content/Proposed_System.tex
\section{Proposed System}
\label{Proposed_system}

This section gives an overview of our user identification pipeline including the used datasets, data pre-processing and segmentation methods, feature extraction, Machine Learning (ML) classifier and accuracy metric used in this study.

\subsection{Datasets}
Four datasets with different stimuli are used: Bioeye TEX, Bioeye RAN, Visual Searching Task, and Gaze on Faces.
See \autoref{datasets_overview} for an overview.

\begin{table}[!ht]
	\caption{Overview of Datasets.}
		\label{datasets_overview}
	\centering
	\renewcommand{\arraystretch}{1.2}
	\begin{tabular}{p{1cm}cccccc}
		\toprule
		\multirow{2}{1.2cm}{Dataset} & \multicolumn{3}{c}{Participants} & Age & Trajectory & Blink\\
		\cline{2-4}
		& M & F & T & Range & Length [s] & Info.\\
		\hline
		TEX &N.A. & N.A.  & 153 & 18--46 & 60 & Y\\		
		RAN & N.A. & N.A. & 153 & 18--46 & 100 & Y \\
		VST & 24 & 34 & 58 &21--33 & 180 & Y\\
		GOF & 193 & 185 & 378 & 20--72 & 60 & N \\
		\bottomrule
	\end{tabular}
\end{table}

\subsubsection{Bioeye (TEX/RAN)}
Two datasets with different stimuli were used in the \emph{BioEye 2015 competition}~\cite{DBLP:journals/ivc/RigasK17}
\footnote{The data was provided by Oleg Komogortsev.}.
Each has two recordings per participant, which were separated by a pause of 30~minutes in between.
\begin{itemize}
	\item[\bf{TEX}] 60~second recordings of reading a poem.
	\autoref{datasetsStimuli:b} shows the gaze trajectory of a sample participant from this dataset along with the reading stimulus.
	\item[\bf{RAN}] 100~second recordings of observing a randomly moving dot.
	\autoref{datasetsStimuli:d} visualizes the gaze trajectory of a sample participant.
\end{itemize}
Both datasets consists of eye movement data from a total of 153 participants including males and females from the age group 18 to 46.

The participants were seated from the monitor screen at a distance of 550 mm.
The dimensions and resolution of the monitor were $474 \times 297$\,mm and $1680 \times 1050$ pixels respectively.
The participant's heads were supported with a chin rest to ensure stability during the sessions to avoid potential eye-tracking artifacts that stem from notable head movements.
The device used for recording was an EyeLink-1000 eye-tracker (1000\,Hz) but the data was down sampled to 250\,Hz using an anti-aliasing FIR filter which interpolated between the invalid gaze points~\cite{DBLP:journals/ivc/RigasK17}.
Despite this interpolation, the dataset still provides the explicit information about the validity of the samples in the recording which can be attributed to device specific faults or user specific reasons (e.g. blinking, loss of attention etc.).
\subsubsection{Visual Searching Task (VST)}
\begin{figure}[ht] 
\begin{subfigure}[b]{0.48\linewidth}
\centering
\includegraphics[width=0.90\linewidth,height=0.20\textheight]{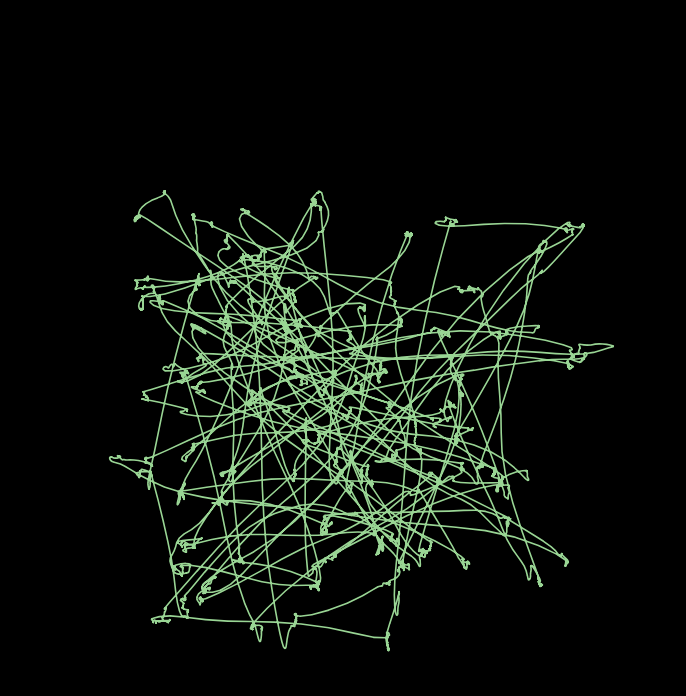} 
\caption{RAN} 
\label{datasetsStimuli:d} 
\end{subfigure} 
\begin{subfigure}[b]{0.48\linewidth}
\centering
\includegraphics[width=0.90\linewidth,height=0.20\textheight]{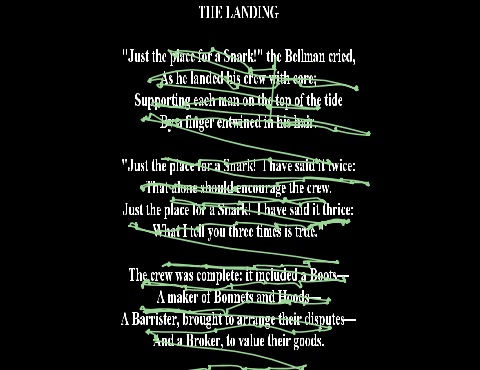} 
\caption{TEX} 
\label{datasetsStimuli:b} 
\end{subfigure} 
\begin{subfigure}[b]{0.5\linewidth}
\centering
\includegraphics[width=0.90\linewidth,height=0.22\textheight]{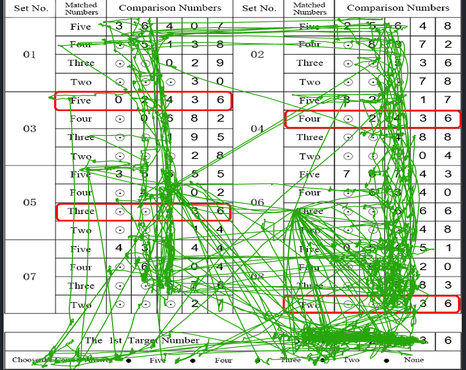} 
\caption{VST} 
\label{datasetsStimuli:a} 
\end{subfigure}
\begin{subfigure}[b]{0.5\linewidth}
\centering
\includegraphics[width=0.86\linewidth,height=0.22\textheight]{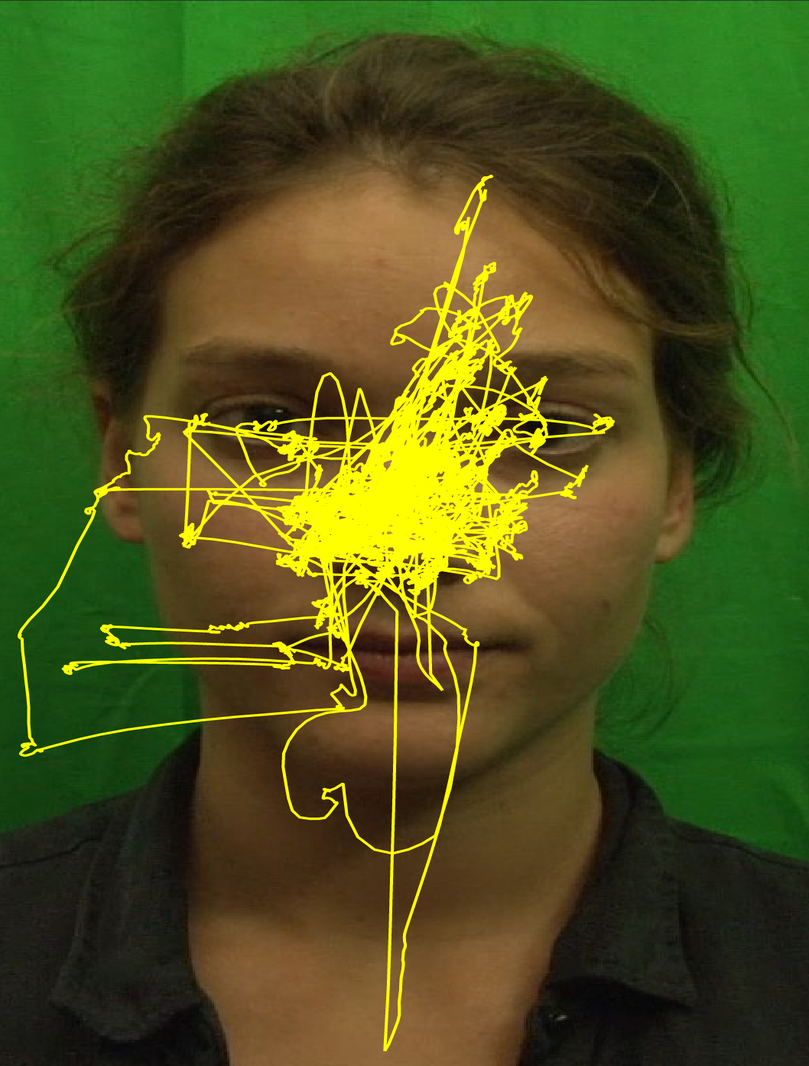} 
\caption{GOF} 
\label{datasetsStimuli:c} 
\end{subfigure}
\caption{Various stimuli and their gaze trajectories.
	(a) A participant looking at random dots from the RAN dataset
	(b) A participant reading a text from the TEX dataset,
	(c) A participant searching for number,
	(d) A participant gazing on face image.}
\label{datasetsStimuli} 
\end{figure}

This dataset~\cite{li2018biometrictask} includes the recording of gaze trajectories of a total of 58 participants (24 males, 34 females) aged 21--33.
The participants performed a visual search task which is a series of number search questions carried out with pictures.
The participants were asked to compare the target number with the comparison numbers in a form to find the longest matched number.
\autoref{datasetsStimuli:a} shows an example of this stimulus and the gaze trajectory of one of the participants.
The recording duration is at least 4 minutes for each participant in each session.
The data collection experiment was divided into two trials with at least two weeks between.
In each trial there were 160 questions divided into four sessions separated by two minutes, each consisting of 40 questions.
Participants took a two minutes rest between these tests.
Overall, $160 \times 2=320$ gaze trajectories for each participant were collected.

The participants were seated from the monitor screen at a distance of 600\,mm.
The dimensions and resolution of the monitor were $474 \times 297$\,mm and $1920 \times 1080$ pixels respectively.
The device used for recording was an Tobii TX300 eye tracking system running at 300\,Hz.
In contrast to the Bioeye dataset, no explicit information about the validity of the gaze trajectory was provided.
Nevertheless, the gaze trajectory consisted of NaNs which we consider as invalid and hence a source for blink information.
We interpolated across the invalid segments in order to have a connected gaze trajectory.

\subsubsection{Gaze on Faces (GOF)}
This dataset is provided in the study of~\cite{coutrot2016face} and it involves the recording gaze trajectories of participants while they observe faces.
The participants looked at multiple images of a single actor gazing towards them for varying durations ($0.1\,s$ to $10.3\,s$) in 32 different trials.
Overall, 8 actors were cast for the videos comprising four females and four males.
Originally the dataset consisted of a total number of 405 candidates from 58 countries, varying in age from 18 to 72.
However, participants that were found with erratic, absent, or damaged data were eliminated to avoid inaccuracies from the study.
Hence, the research in this paper was carried out using the remaining 378 participants of which 193 were males and 185 were females, ranging from ages 20 to 72.

The participants were seated at a distance of 570 mm from a monitor ($1280 \times 1024$ pixels).
The width and height of the stimuli were $429 \times 720$ pixels.
Eye-tracking data was collected using the EyeLink 1000 kit eye-tracker at 250 Hz.~\autoref{datasetsStimuli:c} shows the gaze trajectory of a participant while looking at the face of an actress.
The selected dataset is utilized as it offers an opportunity to study a set of participants that contains a large number of male and female participants with a broader age range.
In this dataset, no invalid parts in the gaze trajectories were found and hence no blink information could be deduced.

\subsection{Preprocessing and Segmentation}
\label{sec_preprocessing_segmentation}
While VST and GOF datasets provide the recorded pixel coordinates directly, RAN and TEX datasets provide viewing angles w.r.t. the x and y axes.
We convert them to screen pixel coordinates $(x_{screen}, y_{screen})$ as follows:

\begin{equation*}
\{x|y\}_{screen} = \left(\frac{ d\times \{w|h\}_{pix}}{\{w|h\}} \right)
tan(\theta_{\{x|y\}}) + \frac{\{w|h\}_{pix}}{2} .
\end{equation*}
The distance from the screen and viewing angles in the $x$ and $y$ directions are denoted by $d$, $\theta_x$, and $\theta_y$ respectively.
$w_{pix}$ and $h_{pix}$ indicate the screen's resolution and  $w$ and $h$ the physical size (width and height). The inverse of the above equation is used to compute the viewing angle coordinates from the pixel coordinates for VST and GOF datasets. 

 
Noise may be present in raw gaze trajectory data which amplifies further in the calculation of velocity, acceleration and other higher order time derivatives of the gaze trajectory.
We implemented a Savitzky-Golay~\cite{Savitzky1964} filter to reduce the noise (see also~\cite{DBLP:journals/spm/Schafer11}).
This filter applies a symmetric polynomial over several points.
In our work we used the polynomial order of 6 and a frame size of 15 as used in the works of ~\cite{george2016score,DBLP:conf/chi/SchroderZPMZ20}.

The Velocity Threshold (IVT) algorithm, utilized in 
numerous publications, see e.g.~\cite{DBLP:conf/icb/HollandK11,DBLP:conf/btas/HollandK12,DBLP:conf/etra/OlsenM12}, 
is used to segment the filtered gaze trajectories into a sequence of fixations and saccades.
The IVT algorithm has been described in different ways in the literature.
In~\cite{sen1984effects,DBLP:conf/etra/SalvucciG00,Andersson2017}, 
the IVT algorithm is implemented with only one parameter i.e. velocity threshold (VT).
This segmentation might produce very short fixations.
These very short fixations are often not meaningful 
because the
brain requires some time to register the visual input~\cite{olsen2012tobii}.
Therefore, many researchers remove the short fixations by using a second parameter named Minimum fixation duration (MFD).
Hence, in various studies~\cite{DBLP:conf/btas/HollandK12,rakoczi2013visual,kasneci2021your}, this algorithm uses both the VT 
and MFD.
We use the IVT algorithm as described in~\cite{george2016score}.
The algorithm defines as fixation all consecutive gaze points resulting in 
eye rotation velocities below the VT, unless the fixation would be shorter than the MFD.
All other segments are identified as saccades.
Commonly used parameters in IVT algorithm include VT = 50\,°/s
and MFD = 100\,ms~\cite{george2016score, DBLP:conf/chi/SchroderZPMZ20}.

The eye tracking data can have invalid data (NaNs) or outliers (e.g invalid gaze point).
This can be due to user-specific reasons such as blinking, loss of attention (micro-sleeping) or eye tracker faults (e.g. solo missed gaze points)~\cite{DBLP:journals/ivc/RigasK17}.
However, the majority of the outliers are reasoned by blinks.
Physiologically, the blinking behavior can encode some information about the participants~\cite{kroger2019does}.
Actual blinking rates vary by individual averaging around 10 blinks per minute and the duration of a blink is on average between 100--400 ms according to the Harvard Database of Useful Biological Numbers\footnote{\url{https://bionumbers.hms.harvard.edu/} [accessed 11-August-2021]}.
Blinking behavior is arguably different in men and women~\cite{doughty2002further} and it has been found that adults blink more often than infants~\cite{blinkINinfantLess}.
These studies motivate us to extract the blink information of the participants in addition to our fixations and saccades when available (which is the case of RAN, TEX and VST datasets).
We extracted the blink segments from the explicitly labeled invalid data (in case of RAN \& TEX) and NaN segments (in case of VST) using a duration threshold between 80–500\,ms, since duration of more than 500\,ms are considered as micro-sleeping~\cite{schleicher2008blinks,wang2011blink} and less than 80 ms can be device faults or other unknown reasons.

\subsection{Feature Extraction}
\label{sec_feature_extraction}
Feature extraction is a basic way to reduce the dimension of high-dimensional data.
For each fixation, saccade and blink (when available), various features are extracted separately.

Let \(X\) and \(Y\) denote the sequences of gaze coordinates in each fixation/saccade where \(X = {x_1, x_2, ..., x_N}\) and \(Y = {y_1, y_2, ..., y_N}\), and \(N\) is the number of points.

We compute a number of basic features, such as 
duration, path length, fix/sac ratio (ratio of maximum fix/sac angular velocity to fix/sac duration), 
fix/sac angle, amplitude, dispersion, distance with the centroid of previous fix/sac, 
angle with the centroid of previous fix/sac and average velocity are computed.
All these features are used in~\cite{george2016score} and the corresponding formulae are
given in that paper. 

\begin{figure*}[!ht]
	\centering
	
	\input{figure/higher_derivative_plots/HOD_formulas.tex}

	\caption{Computation of higher order derivative features.}
	\label{H_O_D_formulas}
\end{figure*}
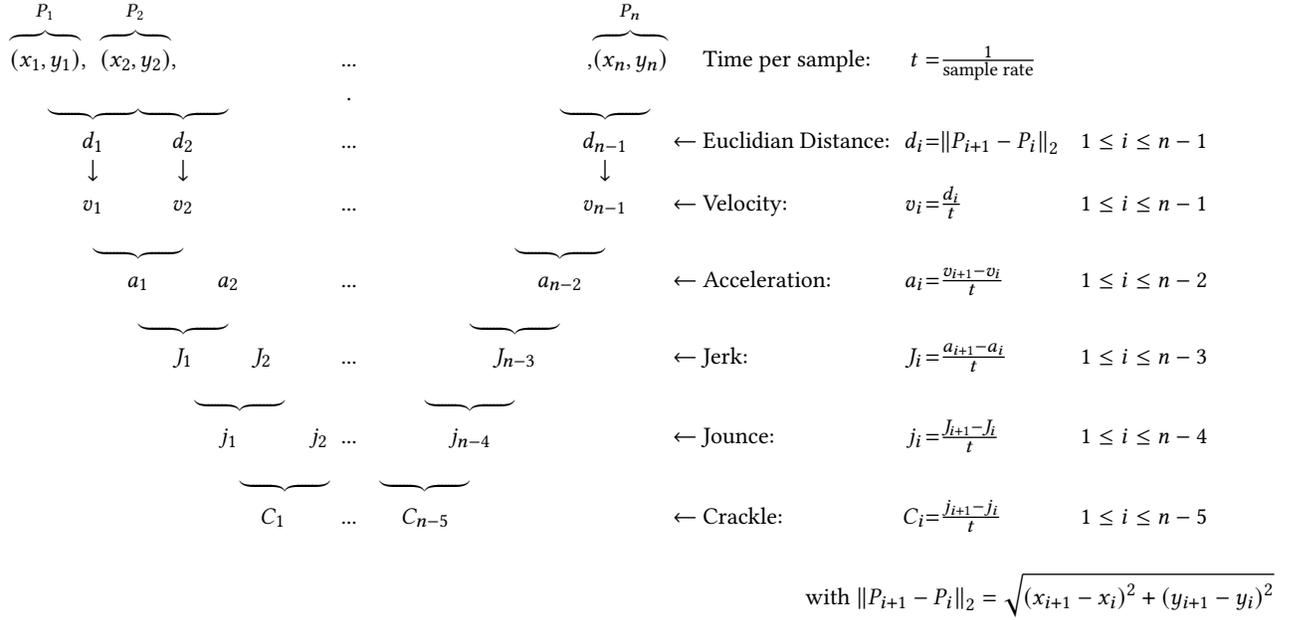

All the derivatives, such as velocity, acceleration, jerk, etc. are computed using the ``forward difference method'' as detailed in \autoref{H_O_D_formulas}.

Since for $k\geq 1$, the $k$-th order derivative can only be computed for segments (fixations or saccades) of at least $(k+1)$ points. 
We exclude, whenever we use such higher-order derivatves, segments that are shorter than $(k+1)$ points (by appropriately merging the neighboring segments, in accordance to our IVT algorithm).
We always remove saccades consisting of only one or two points. 
As shown in \autoref{user_prediction_features}, various statistical features namely mean, median, max, standard deviation, skewness and 
kurtosis are computed for different types of velocities and derivatives (in the following refereed as M3S2K features).

To the best of our knowledge, previous approaches to eye movement biometrics only include derivaties
until order two, i.e., they never use derivatives beyond that of acceleration.
In this paper, we demonstrate that including higher order derivatives beyond acceleration
can indeed be beneficial for eye movement biometrics.
There are works within eye tracking reserach that include up to fourth order derivatives 
(to predict saccade movements, see~\cite{wang2017dynamic}).
In general, it is well known that humans'
predictive capabilities in their perception-action loop can
be capture by higher order derivatives of the perception or 
action trajectories~\cite{sargolzaei2016sensorimotor}.

Lastly, seven features namely number of blinks, duration of each blink, total of duration, mean of duration, minimum of duration, maximum of duration, and variance of duration are computed from the blink segments as shown in~\autoref{user_prediction_blink_features}.
We tried on blink duration with different sets of the M3S2K statistical features, and these seven features gave the best results.

In all cases, the features are normalized using the Z-score standardization/normalization method implemented by \cite{scikit-learn} (sklearn.\allowbreak preprocessing.\allowbreak StandardScaler).
The method of calculation here is to determine the mean and standard deviation for each feature.
Next we subtract the mean from each feature and divide the obtained value by its standard deviation.
This ensures that each feature's values are in the similar range (all features are centered around 0 and have variance in the same order) and therefore contribute equally to the classification.

\begin{table}[ht]
	\centering
	\caption{User identification features.}
	\label{user_prediction_features}
	\begingroup
	\setlength{\tabcolsep}{2.5pt} 
	\renewcommand{\arraystretch}{1} 
	\begin{tabular}{ll|ll}
		\toprule
		{} & \begin{tabular}{@{}c@{}}Fix./Sac.\\Features\end{tabular} & {}  & \begin{tabular}{@{}c@{}}Fix./Sac.\\Features\end{tabular}\\
		\midrule
		1& Duration & 16--21 & Angular velocity* \\
		2& Path length & 22--27& Velocity X* \\
		3& Skew X & 28--33&  Velocity Y* \\
		4& Skew Y & 34--39& Angular acceleration* \\
		5& Kurt X & 40--45& Acceleration X*  \\
		6& Kurt Y & 46--51& Acceleration Y* \\
		7& STD of X & 52--57&  Angular jerk*  \\
		8& STD of Y & 58--63&  Jerk X* \\
		9& Fix/Sac ratio  & 64--69&  Jerk Y* \\
		10& Fix/Sac angle & 70--75 & Angular jounce* \\
		11& Amplitude & 76--81& Jounce X* \\
		12& Dispersion & 82--87 & Jounce Y* \\
		13& Dist. with previous Fix/Sac& 88--93 & Angular crackle* \\
		14& Angle with previous Fix/Sac& 94--99 & Crackle X* \\
		15& Average velocity & 100--105 & Crackle Y*  \\
		\bottomrule
	\end{tabular}
	\endgroup
	\raggedright
	
	\vspace{1mm}
	*M3S2K-Statistical features:\\
	\phantom{*}Mean, Median, Max, Std, Skewness, Kurtosis
\end{table}

\begin{table}[ht]
	\centering
	\caption{Blinks features.}
	\label{user_prediction_blink_features}
	\begingroup
	\setlength{\tabcolsep}{6pt} 
	\renewcommand{\arraystretch}{1} 
	\begin{tabular}{ll|ll}
		\toprule
		{} & Blinks Features & {}  & Blinks Features\\
		\midrule
		1& Duration & 5 & Minimum of the duration \\
		2& Number of blinks & 6& Maximum of the duration\\
		3& Mean of the duration & 7& Variance of the duration \\
		4& Total of the duration & &  \\
		\bottomrule
	\end{tabular}
	\endgroup
	\raggedright
\end{table}

\subsection{Machine Learning Classifier}
\label{ML_classifiers}
For given trajectories of eye movements of a number of participants, we seek an algorithm that, given an unseen trajectory, is able to detect which participant has generated it.
This type of problem is known as a \emph{classification problem}, the algorithm that carries out the classification is known as \emph{classifier}.
The machine learning classifier Radial Basis Function Networks (RBFN)~\cite{broomhead1988radial} has been used in~\cite{george2016score} and also the work in~\cite{DBLP:conf/chi/SchroderZPMZ20}.
We compared RBFN with Random Forest~(RF)~\cite{DBLP:journals/ml/Breiman01} and we found that RBFN gave better accuracies.
Therefore, we use RBFN in this paper.
A maximum of three instances of the RBFN classifiers are trained: one to predict the user from fixation segments, the second to predict the user from saccade segments and third to predict user from blink segments (whenever available).
The final prediction probability $p^i_\text{final}$ is the weighted average of the probabilities of fixation ($p^i_\text{fix}$), saccade ($p^i_\text{sac}$) and blink ($p^i_\text{blink}$) classifiers for each class $i$ (user ID):
\begin{equation*}
\label{final_Prob}
p^i_\text{final} = p^i_\text{fix} w_\text{fix} + p^i_\text{sac} w_\text{sac} + p^i_\text{blink} w_\text{blink} ,
\end{equation*}
where $w_\text{fix}$, $w_\text{sac}$, and $w_\text{blink}$ are the weights for the fixation, saccade and blink classifiers respectively.
In case the blink classifier is absent, $w_\text{fix}$, $w_\text{sac}$ are typically selected as 0.5 each~\cite{DBLP:conf/chi/SchroderZPMZ20, george2016score}.
For $m$ classes, the class having the maximum probability $p_{\text{max}} = \max \{p^i_\text{final} \mid i \in (1,\ldots,m) \} $ is final outcome of the ML classifier.

\subsection{Performance Metrics}
\label{Performance_Metrics_}
In our case a class is a user of an eye tracking device in one of the experiments.
For a single experiment, we always calculate predictions for all available users.
As accuracy of user identification we divide the number of correct predictions by the total number of predictions (equal to the number of users).
Since the result of the RBFN algorithm depends randomly on the initialization of its internal state, 
we perform a cross-validation with 50 different states (seeds) and take the average accuracy as the final accuracy of user identification for each experiment in this work.
Together with the final accuracy, we also report the \emph{standard error of the mean}:
$\sigma_\mu = \frac{\sigma}{\sqrt{k}}$,
where $\sigma$ is the standard deviation of the results and $k$ the number of runs.

%% file: figure/higher_derivative_plots/HOD_formulas.tex

\newlength{\tabcolsepbefore}
\setlength{\tabcolsepbefore}{\tabcolsep}
\setlength{\tabcolsep}{0mm}

\newcommand{\tsh}{3mm}
\newcommand{\tsf}{6mm}
\newcommand{\tsd}{12mm}

\newcolumntype{C}[1]{>{\centering\arraybackslash}p{#1}}
	
\begin{tabular}{
		C{\tsh} C{\tsh} C{\tsh} 
		C{\tsh} C{\tsh} 
		C{\tsh} C{\tsh} 
		C{\tsh} C{\tsh} 
		C{\tsh} C{\tsh} 
		C{\tsf} C{\tsf} 
		c 
		C{\tsf} C{\tsf} C{\tsf} C{\tsf} C{\tsf} C{\tsf} C{\tsf} 
		l l c c l c 
	 }
 
	
	\multicolumn{4}{c}{$\overbrace{\left(x_1, y_1\right)}^{P_1}$,} & \multicolumn{4}{c}{$\overbrace{\left(x_2, y_2\right)}^{P_2}$,} & & & & & & ...
 	& & & & & &  \multicolumn{2}{c}{,$\overbrace{\left(x_n,y_n\right)}^{P_n}$} & \hspace{4mm} & Time per sample: & $t$ & $=$ & $\frac{1}{\textrm{sample rate}}$ & \\
	& & \multicolumn{4}{c}{$\underbrace{\hspace{\tsd}}$} & \multicolumn{4}{c}{$\underbrace{\hspace{\tsd}}$} & & & & .
	& & & & & \multicolumn{3}{c}{$\underbrace{\hspace{\tsd}}$} \\
	& & & \multicolumn{2}{c}{${d_1}$} & & & \multicolumn{2}{c}{${d_2}$} & & & & & ...
	& & & & & & $d_{n-1}$ & & $\leftarrow$ & Euclidian Distance:  \hspace{1mm} & $d_i$ & $=$ & $\left\| P_{i+1} - P_{i} \right\|_2$ \hspace{2mm} & $1\leq i\leq n-1$ \\
	& & & \multicolumn{2}{c}{$\downarrow$} & & & \multicolumn{2}{c}{$\downarrow$} & & & & &
	& & & & & & $\downarrow$ \\
	& & & \multicolumn{2}{c}{${v_1}$} & & & \multicolumn{2}{c}{${v_2}$} & & & & & ...
	& & & & & & $v_{n-1}$ & & $\leftarrow$ & Velocity: & $v_i$ & $=$ & $\frac{d_i}{t}$ & $1\leq i\leq n-1$ \\
	& & & & \multicolumn{4}{c}{$\underbrace{\hspace{\tsd}}$} & & & & & &
	& & & & \multicolumn{3}{c}{$\underbrace{\hspace{\tsd}}$} \\
	& & & & & \multicolumn{2}{c}{${a_1}$} & & & \multicolumn{2}{c}{${a_2}$} & & & ...
	& & & & & $a_{n-2}$ & & & $\leftarrow$ & Acceleration: & $a_i$ & $=$ & $\frac{v_{i+1}-v_i}{t}$ & $1\leq i\leq n-2$ \\
	& & & & & & \multicolumn{4}{c}{$\underbrace{\hspace{\tsd}}$} & & & &
	& & & \multicolumn{3}{c}{$\underbrace{\hspace{\tsd}}$} \\
	& & & & & & & \multicolumn{2}{c}{$J_1$} & & \multicolumn{2}{c}{$J_2$} & & ...
	& & & & $J_{n-3}$ & & & & $\leftarrow$ & Jerk: & $J_i$ & $=$ & $\frac{a_{i+1}-a_i}{t}$ & $1\leq i\leq n-3$ \\
	& & & & & & & & \multicolumn{4}{c}{$\underbrace{\hspace{\tsd}}$} & &
	& & \multicolumn{3}{c}{$\underbrace{\hspace{\tsd}}$}\\
	& & & & & & & & & \multicolumn{2}{c}{$j_1$} & & $j_2$ & ...
	& & & $j_{n-4}$ & & & & & $\leftarrow$ & Jounce: & $j_i$ & $=$ & $\frac{J_{i+1}-J_i}{t}$ & $1\leq i\leq n-4$ \\
	& & & & & & & & & & \multicolumn{3}{c}{$\underbrace{\hspace{\tsd}}$} &
	& \multicolumn{3}{c}{$\underbrace{\hspace{\tsd}}$} \\
	& & & & & & & & & & & $C_1$ & & ...
	& & $C_{n-5}$ & & & & & & $\leftarrow$ & Crackle: & $C_i$ & $=$ & $\frac{j_{i+1}-j_i}{t}$ & $1\leq i\leq n-5$ \\
	
\end{tabular}

\vspace{5mm}
\hfill with  $\left\| P_{i+1} - P_{i} \right\|_2 = \sqrt{\left(x_{i+1}-x_i\right)^2+\left(y_{i+1}-y_i\right)^2}$

\setlength{\tabcolsep}{\tabcolsepbefore}

%% file: content/User_Prediction_Experiments.tex
\section{User Identification Experiments}
\label{sec_Experiments}

This section introduces different experiments that are employed to investigate 
the individual effects towards prediction accuracy of 
stimuli, IVT parameters, including higher order speed derivatives, 
including blink features, and including gender and age across the different datasets.

\subsection{Effect of Stimuli}
\label{sec_effect_of_stimuli}
We study the accuracy of user identification in the four different datasets with different stimuli.
We use the default IVT parameters of VT = 50\,°/s, MFD = 100\,ms for the RAN, TEX, and VST datasets.
For GOF VT = 15\,°/s is applied, since the default parameters lead to zero fixations for some participants in this dataset.
We use the first 51 features shown in \autoref{user_prediction_features} and an equal weighting of the saccade and fixation classifiers to compute the user prediction probabilities.
In the following, the user identification experiments are explained and.
The results are summarized in~\autoref{Accuracy_with_51_features_of_all_datasets}.
\begin{table}[ht]
	\caption{Performance metrics with different datasets using 51 features over 50 runs.}
	\label{Accuracy_with_51_features_of_all_datasets}
	\begin{tabular}{cccc}
		\toprule
		\begin{tabular}{@{}c@{}}Data\\set\end{tabular} & \begin{tabular}{@{}c@{}}Identification\\Accuracy\end{tabular} & \begin{tabular}{@{}c@{}}Number of\\ participants\end{tabular}  & \begin{tabular}{@{}c@{}}Trajectory\\ Length [s]\end{tabular}\\
		\midrule
		RAN & 92.62 $\pm$ 0.13\,\% & 153  & 100\\
		TEX & 90.90 $\pm$ 0.10\,\% & 153  & 60 \\
		VST & 85.69 $\pm$ 0.16\,\% & 58  & 180\\
		GOF & 77.91 $\pm$ 0.51\,\% & 153  & 60\\
		\bottomrule
	\end{tabular}
\end{table}
\paragraph{Bioeye TEX/RAN}
\label{user_prediction_bioeye_data}
For both datasets, the ML classifier is trained with all the 153 participants of the second session and tested with the first session.
In our previous work we achieved an identification accuracy of 94.10\,\% using RAN data and 90.80\,\% using TEX data with \textbf{one run}~\cite{DBLP:conf/chi/SchroderZPMZ20}.
In this work, the average accuracy achieved over \textbf{50 runs} with the RAN dataset is 92.62 $\pm$ 0.13\,\% (maximum accuracy = 94.77\,\%) and with TEX dataset is 90.90 $\pm$ 0.10\,\% (maximum accuracy = 92.28\,\%) which is a more stable prediction accuracy of our classifier compared to our previously reported results in~\cite{DBLP:conf/chi/SchroderZPMZ20}. 

\paragraph{VST}
\label{user_prediction_VSTe_data}
%
The ML classifier is trained with all the 58 participants and use training and testing sessions that are recorded in the same day.
The accuracy achieved with this data set is 85.69 $\pm$ 0.16\,\%.

\paragraph{GOF}
\label{user_prediction_GOF_data}
This dataset has a large number of participants (193 males and 185 females in different age groups).
We used the 
first 16 trials for training and the 
remaining 16 trials for testing.
For better comparison with BioEye data, 153 participants (77 males and 76 females) are chosen randomly over 50 runs and the average accuracy is 77.91 $\pm$ 0.51\,\%.


\subsection{Effect of IVT Parameters}
\label{IVT_parameters}
As mentioned previously, the IVT algorithm has two parameters namely velocity threshold (VT) and minimum fixation duration (MFD).
In order to study the effect of changing the IVT parameters, a systematic parameter variation is conducted to determine which IVT parameter leads to the highest accuracy of our ML classifier.

\begin{figure}[ht] 
	\begin{subfigure}[b]{1\linewidth}
		\centering
		\includegraphics[width=1\linewidth]{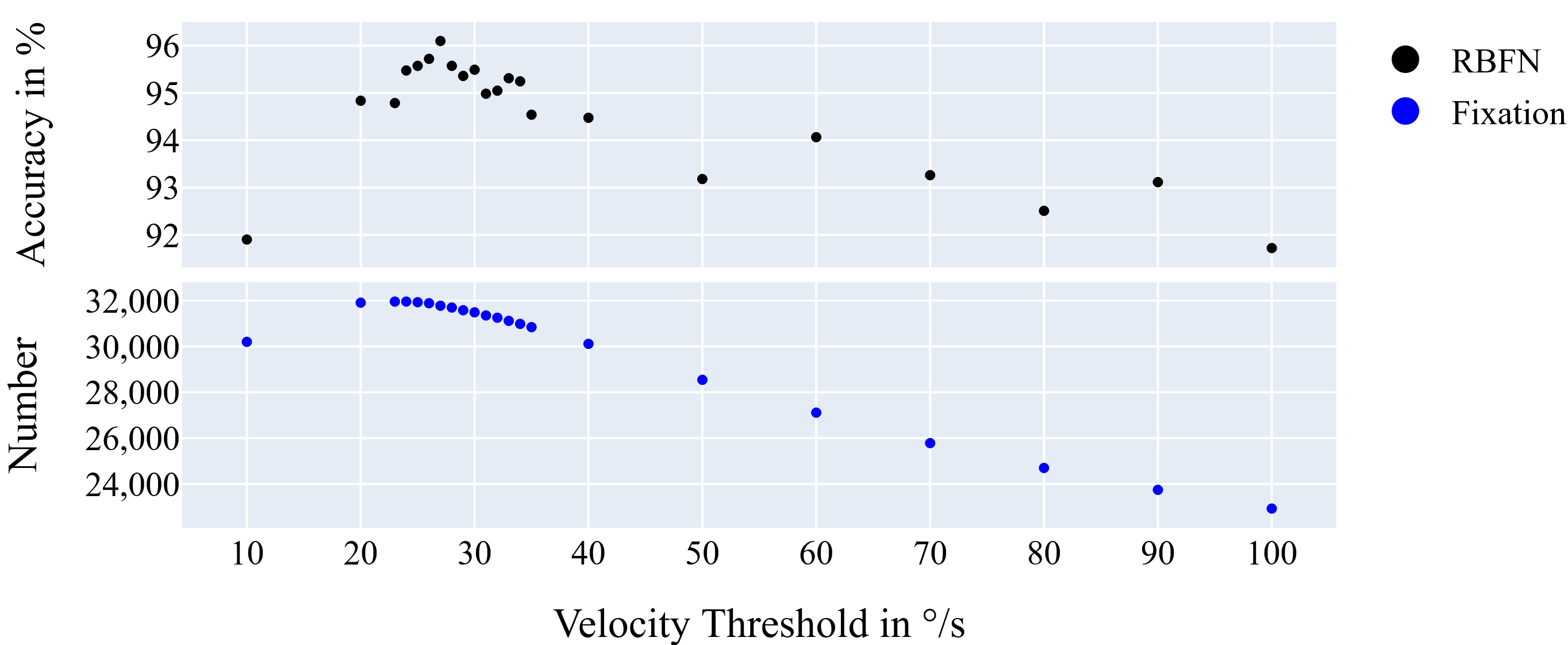} 
		\caption{RAN} 
		\label{fig:ivt-parametersRAN} 
		\vspace{4ex}
	\end{subfigure} 
	\begin{subfigure}[b]{1\linewidth}
		\centering
		\includegraphics[width=1\linewidth]{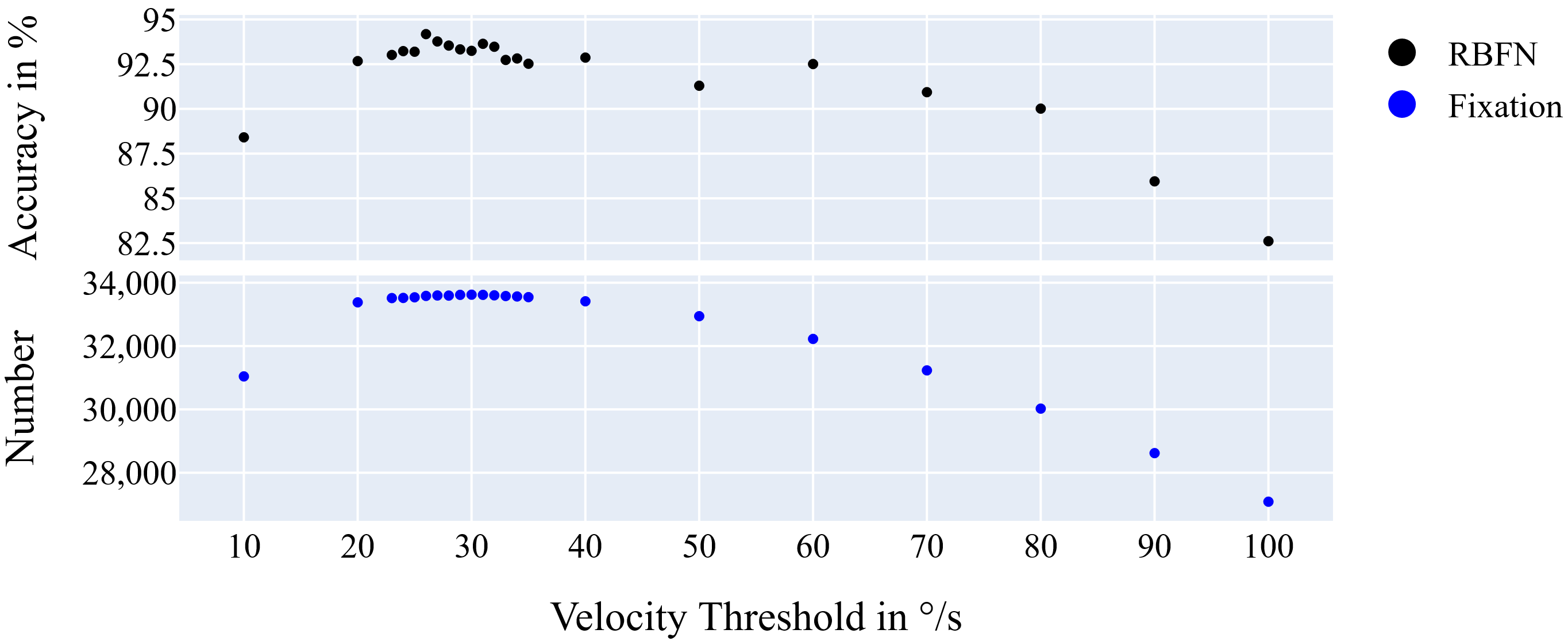} 
		\caption{TEX} 
		\label{fig:ivt-parametersTEX} 
	\end{subfigure} 
	\caption{Variation of Velocity thresholds of BioEye Data with fixing MFD at 100\,ms.}
	\label{fig:ivt-parameters} 
\end{figure}

An initial set of experiment is carried out with the RAN and TEX datasets.
In the first stage, we vary the VT with a fixed MFD of 100\,ms and in the second stage we fix the VT (at the value with the highest accuracy from the previous stage) and vary the MFD.
The parameter range under consideration was 10--100\,°/s for VT and 50--150\,ms for MFD.
In each stage, first a broad variation is done in steps of 10, then a fine variation is executed in steps of 1.
For each setting, we perform a cross validation with random 80\,\%
subsets of the users (i.e. 122 participants) for a total of 50 runs.
\autoref{fig:ivt-parametersRAN} and~\autoref{fig:ivt-parametersTEX} show the accuracy of user identification along with the number of fixations as the velocity threshold is varied.
It can be noted that the highest accuracy is always obtained around the highest number of fixations.
The best accuracy is achieved with a VT and MFD respectively of 26 °/s and 98\,ms for TEX, 27 °/s and 96 ms for RAN.
With MFD = 100 ms the obtained accuracies are almost identical (the difference is in the order of 0.05\,\%).

\begin{figure}
	\centering
	\includegraphics[width=0.99\linewidth]{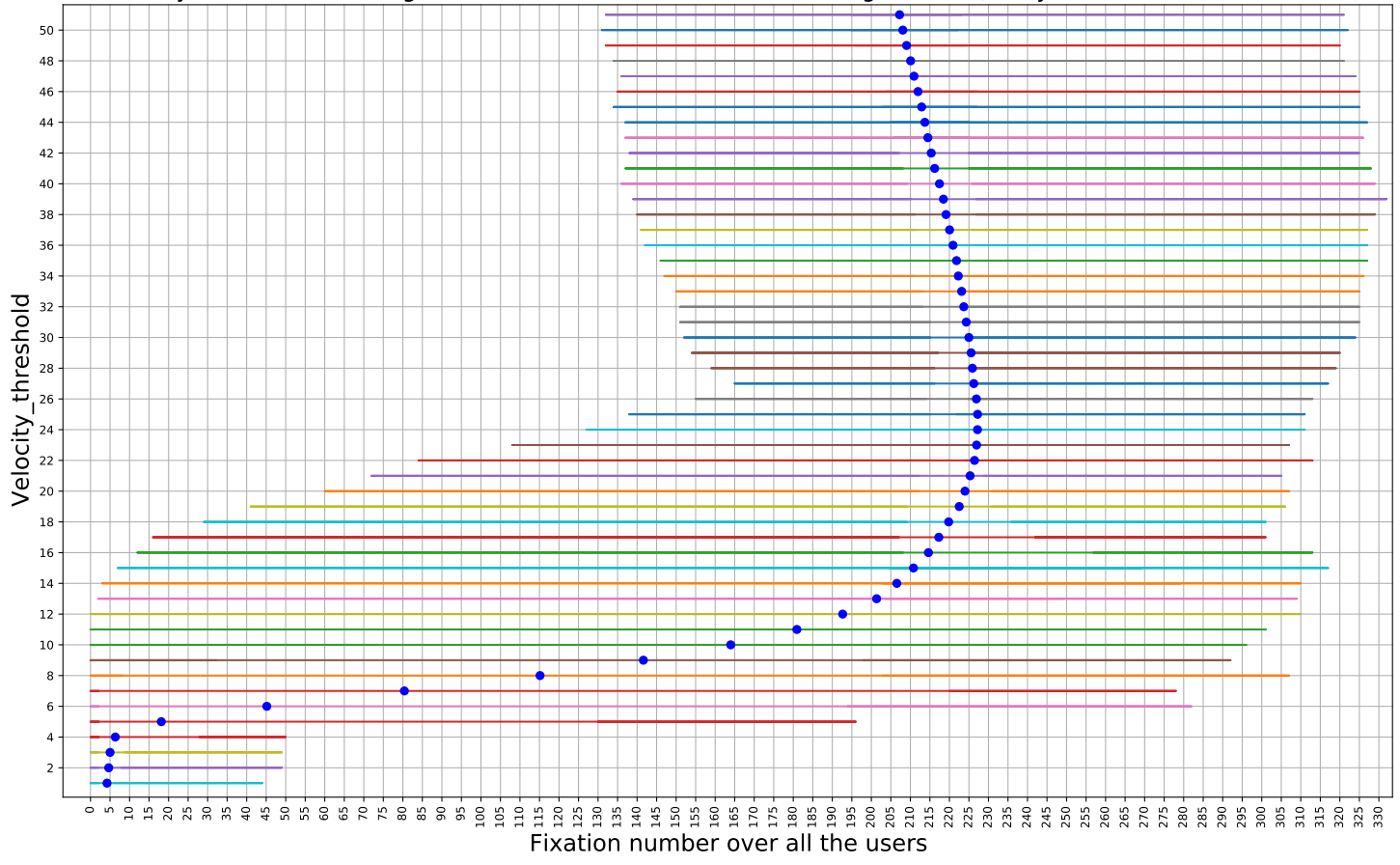}
	\caption{Velocity threshold of IVT against number of fixation for training session (RAN data set / 153 participants).}
	\label{fig:ivt-RAN-Zoomed}
\end{figure}

The above approach requires a cross validation in the initial stages to compute the 
best parameters which is a time consuming process.
Therefore, in an alternative approach we first tune the VT parameter to achieve the highest number of fixation.
This is done as follows. We first prepare a plot such as~\autoref{fig:ivt-RAN-Zoomed} 
(which is for the RAN dataset).
The $x$-coordinate represents the number of fixations over all the participants in the data set and the $y$-coordinate shows the velocity thresholds.
The colored stripes are a representation of the number of fixations of all the participants in the data set for each velocity threshold.
The beginning of the line indicates the minimum number of fixations, while the end of the line indicates the maximum number of fixations on each specific velocity threshold.
The blue dots on the colored stripes mark the average fixation number.
From the plot we can determine a peak in the number of fixations at a VT value of approximately $24$.
We then determine the precise accuracies for this VT value, plus a few neighboring VT values and choose the VT value that gives the highest accuracy within this range.

We performed the above described procedure and obtained the following results for the four data sets using 51 features in fixation and saccade classifiers:
\begin{itemize}
	\item[\bf RAN] The highest number of fixations in RAN data set occurred with a velocity threshold of 24\,°/s with accuracy of 94.89\,\%, while the best accuracy of 95.96\,\% is achieved with a VT of 27\,°/s and MFD of 100 ms over 50 runs.
	\item[\bf TEX] The best accuracy of 93.23\,\% is achieved with a VT of 26\,°/s  and MFD of 100\,ms, while the highest number of fixations occurred with a VT of 30\,°/s with accuracy of 92.24\,\% using TEX data set.
	\item[\bf VST] The best accuracy of 94.82\,\% is attained with a VT of 100\,°/s, MFD of 100\,ms and the highest number of fixation occurred at a VT of 120\,°/s with accuracy of 94.31\,\%.
	\item[\bf GOF] In this data set, The VT of 21\,°/s produced the highest number of fixations with accuracy of 82.35\,\% and the best accuracy of 83.45\,\% is achieved with the VT of 22\,°/s  and MFD of 100\,ms over 152 participants (76 males and 76 females) which are selected randomly over 50 runs from all the data users.
\end{itemize}
In all the above cases, we were able to significantly increase the accuracies of user identification (by 3.34\,\% for RAN, 2.33\,\% for TEX, 9.03\,\% for VST and 4.76\,\% for GOF) by selecting the optimal IVT parameters (compare with \autoref{Accuracy_with_51_features_of_all_datasets}).

\subsection{Effect of Higher-Order Derivatives}
\label{sec_higher_order_derivatives}
In order to study the effect of higher order derivative features on the accuracy of user identification, we start with a minimal set of first 13 position based features and duration as a general feature.
See the first 14 features in~\autoref{user_prediction_features}.
Next we add 19 velocity based features and than, step by step, we include 18 statistical features based on each of the other higher order derivatives (acceleration, jerk, jounce, crackle).
For RAN, TEX and VST, the IVT VT parameter is chosen as 50\,°/s and for GOF dataset, it is 15\,°/s since the default parameters lead to zero fixations for some participants in this dataset.

\autoref{user_prediction_dervatives_features_all_data} shows the accuracy with an increasing number of higher order derivative features for the four different datasets.
The accuracy of user identification increases until the inclusion of jounce features for the RAN, TEX and GOF datasets.
For the crackle based features, the accuracy decreases again.
For the VST dataset, the accuracy rises only until the jerk level features and decreases already for jounce.


The shown observations attest our hypothesis that including higher order derivatives (up to a certain level) is indeed a useful way to capture meaningful information that contribute to an increase in the accuracy of user identification.
\begin{table*}[!ht]
	\centering
	\caption{Performance metrics over 50 runs with varying number of features of higher order derivatives of the gaze trajectory of all the datasets.}
	\label{user_prediction_dervatives_features_all_data}
	\begingroup
	\setlength{\tabcolsep}{5pt} 
	\renewcommand{\arraystretch}{1.1} 
	\begin{tabular}{clccccc}
		\toprule
		& Derivative &
		\begin{tabular}{@{}c@{}}Number\\of\\Features\end{tabular} &
		\begin{tabular}{@{}c@{}}RAN (153 participants)\\ Acc. VT=50\,°/s,\\ MFD=100\,ms\end{tabular} &
		\begin{tabular}{@{}c@{}}TEX (153 participants)\\ Acc. VT=50\,°/s,\\ MFD=100\,ms\end{tabular} &
		\begin{tabular}{@{}c@{}}VST (58 participants)\\ Acc. VT=50\,°/s,\\ MFD=100\,ms\end{tabular} &
		\begin{tabular}{@{}c@{}}GOF (153 participants)\\ Acc. VT=15\,°/s,\\ MFD=100\,ms\end{tabular}\\
		\midrule
		0. & Position & 14 & 83.02 $\pm$ 0.22\,\%& 84.30 $\pm$ 0.17\,\% & 84.79 $\pm$ 0.22\,\%& 46.82 $\pm$ 0.44\,\%\\
		1. & Velocity & 33 &  89.67 $\pm$ 0.15\,\% & 89.81 $\pm$ 0.11\,\% & 85.03 $\pm$ 0.15\,\%& 69.46 $\pm$ 0.54\,\%\\
		2. & Acceleration & 51 & 92.62 $\pm$ 0.13\,\% & 90.90 $\pm$ 0.10\,\%&85.69$\pm$ 0.16\,\%&77.91 $\pm$ 0.51\,\%\\
		3. & Jerk & 69 & 93.54 $\pm$ 0.10\,\%  & 91.73 $\pm$ 0.12\,\%  &\bf{86.52} $\pm$ 0.23\,\% &80.08 $\pm$ 0.50\,\%\\
		4. & Jounce & 87 & \bf{94.58} $\pm$ 0.10\,\%  &\bf{91.96} $\pm$ 0.08\,\%&85.13 $\pm$ 0.17\,\%&\bf{81.02} $\pm$ 0.53\,\%\\
		5. & Crackle & 105 & 94.15 $\pm$ 0.09\,\% & 90.86 $\pm$ 0.14\,\% & 83.82 $\pm$ 0.14\,\% & 80.54 $\pm$ 0.54\,\%\\
		\bottomrule
	\end{tabular}
	\endgroup
\end{table*}
\subsection{Effect of Blinking Features}
\label{Effect_of_blinking_features}
We study the effect of including the blink classifier as introduced in~\autoref{ML_classifiers} in addition to fixation and saccade classifiers.
Since, the identification accuracy of the blink classifier is lower than the fixation and saccade classifiers, we can not weigh all of them equally to compute the final accuracy like done before.
In order to find the optimal weights for our three classifiers, we use the Nelder-Mead Method~\cite{1965_nelder_mead} which is a popular direct search method (based on function comparison~\cite{gao2012implementing}) and is suited for optimization problems for which derivatives may not be known.
As shown in~\autoref{user_prediction_with_blink_CLF_bioeye_RAN_TEX_VST}, the accuracy increases by 1.35\,\%, 1.25\,\%, and 0.5\,\% with use of the blink classifier in VST, RAN, and TEX dataset respectively.
\begin{table}[ht]
\centering
\caption{Performance metrics over 50 runs using blink classifier in \bf{RAN, TEX, and VST datasets} (VT = 50\,°/s and MFD = 100\,ms).}
	\label{user_prediction_with_blink_CLF_bioeye_RAN_TEX_VST}
	\begingroup
	\setlength{\tabcolsep}{2pt} 
	\renewcommand{\arraystretch}{1} 
	\begin{tabular}{cccc}
		\toprule
		Dataset&\begin{tabular}{@{}c@{}}Sac/Fix/blink\\features\end{tabular} & \begin{tabular}{@{}c@{}}CLF weights\\ Sac/Fix/Blink\end{tabular} & \begin{tabular}{@{}c@{}}Identification\\ Accuracy \end{tabular} \\
		\midrule
		RAN & 51/51/0 & 0.5/0.5/0.0 & 92.62 $\pm$ 0.13\,\%\\
		& 51/51/7 & 0.408/0.578/0.015 & 93.87 $\pm$ 0.12\,\%\\
		\bottomrule
		TEX& 51/51/0 & 0.5/0.5/0.0 & 90.90 $\pm$ 0.10\,\%\\
		& 51/51/7 & 0.4453/0.5453/0.0094 & 91.40 $\pm$ 0.10\,\%\\
		\bottomrule
		VST& 51/51/0 & 0.5/0.5/0.0 & 85.69 $\pm$ 0.16\,\%\\
		& 51/51/7 & 0.568/0.3938/0.0381 & 87.04 $\pm$ 0.24\,\%\\
\bottomrule
	\end{tabular}
	\endgroup
\end{table}

\subsection{Effect of Gender and Age Groups on User Identification}
We study the effect of gender and age on the accuracy of user identification using the GOF dataset.
It has large number of participants with gender and age information of the participants (\autoref{GOF_Demographics}).
\begin{table}[!ht]
	\centering
	\caption{Demographics of Males and Females in GOF dataset.}
	\label{GOF_Demographics}
	\begin{tabular}{cccc}
		\midrule
		Age Group & Males & Females & Total \\ \hline
		20 -- 40& 151& 157 & 308 \\
		41 -- 72& 42& 28 & 70 \\
		\bottomrule
	\end{tabular}
\end{table}
\subsubsection{Gender}
To study the effect of gender, user identification experiments are performed in three different groups with each 150 participants: a balanced group with 75 Males and 75 Females, a group exclusively with male participants and a group exclusively with female participants.
The participants of the groups were randomly chosen from the complete dataset 50 times each.
The average of the user identification accuracies are reported in~\autoref{GOF_gender_experiments_new}.
For the mixed group, the accuracy achieved is 83.25 $\pm$ 0.48\,\% .
In isolated groups of males and females, the accuracy is found to be higher in the female group (88.85 $\pm$ 0.32\,\%) when compared to the male one (77.37 $\pm$ 0.50\,\%).
This shows that user identification is biased by gender and is found to be more accurate in female than male group of participants.

\subsubsection{Age}
To study the effect of age, the participants were split into two age groups 20--40 years and 41--72 years.
Both age groups contain 56 participants with an equal number of males and females.
The user identification can be performed in the older age group with a five percentage points higher accuracy in comparison to the younger age group (\autoref{GOF_gender_experiments_new}).
\begin{table}[ht]
	\centering
	\caption{Effect of gender and age groups on average user identification accuracy over 50 runs with VT = 22 using 51 features.}
	\label{GOF_gender_experiments_new}
	\begin{tabular}{ccccc}
		\toprule
		Age group & \specialcell{Number\\of participants} & M & F & \specialcell{Identification\\Accuracy} \\
		\midrule
		20 -- 72 & 150 & 75 & 75 & 83.25 $\pm$ 0.48\,\% \\
		20 -- 72 & 150 & 150 & 0 & 77.37 $\pm$ 0.50\,\% \\
		20 -- 72 & 150 & 0 & 150 & 88.85 $\pm$ 0.32\,\% \\
		\midrule
		20 -- 40 & 56 & 28 & 28 & 85.96 $\pm$ 0.79\,\% \\
		41 -- 72 & 56 & 28 & 28  & 91.43 $\pm$ 0.47\,\% \\
		\bottomrule
	\end{tabular}
\end{table}

\subsection{Effect of the Time Gap Between Train and Test Data}
Finally, we study the effect of the time gap between train and test data using the RAN, TEX and VST data sets.
The effect is also referred to as template aging~\cite{george2016score}.

\subsubsection{BioEye datasets}
Out of 153 participants, this dataset has 37 participants for which recordings are also available after one year.
The first experiment in this dataset is conducted with these participants.
The IVT parameters used in these experiments are VT = 50 °/s and MFD = 100 ms.
For the RAN dataset, the average accuracy over \textbf{50 runs} is 83.51 $\pm$ 0.18\,\%
and with TEX the accuracy is 75.24 $\pm$ 0.32\,\%.
In comparison with the work of George and Routray~\cite{george2016score} running the same experiment yields 81.08\,\% and 78.38\,\% with \textbf{one run} for RAN and TEX respectively.

\subsubsection{VST dataset}
This dataset has a time gap of more than two weeks between the two trials and two hours between four sessions in each trial.
The accuracies of user identification were noted as $(94.6 \pm  0.1)\,\%$ (2 hour time gap i.e. train with session 1 and test with session 4 both in trial 1) and $(76.0 \pm 0.2)\,\%$ (2 week time gap i.e. train with session 1, trial 1 and test with session 1, trial 2).

\medskip
As expected, the accuracy of user identification decreases for both datasets when there is a significant time gap between train and test sessions.
This may be attributed to changing physiological parameters of the participants, device characteristics, and some other inexplicable effects.

%% file: content/Discussion.tex
\section{Study of Combined Factors} 
\label{Discussion}

In this section, we combine the effects of above factors which leads to further improvement in user identification accuracy.

\begin{table*}
	\centering
	\caption{Performance metrics over 50 runs with varying number of features of higher order derivatives of the gaze trajectory of all the datasets with their best Velocity Threshold (VT).}
	\label{user_prediction_dervatives_features_all_data_best_VT}
	\begingroup
	\setlength{\tabcolsep}{5pt} 
	\renewcommand{\arraystretch}{1.1} 
	\begin{tabular}{cccccc}
		\toprule
		\begin{tabular}{@{}c@{}}Derivatives\\order\end{tabular} &
		\begin{tabular}{@{}c@{}}Features\\Number\end{tabular} &
		\begin{tabular}{@{}c@{}}RAN (153 participants) \\ Acc. VT=27 °/s,\\ MFD= 96 ms\end{tabular} &
		\begin{tabular}{@{}c@{}}TEX (153 participants) \\ Acc. VT=26 °/s,\\ MFD= 98 ms\end{tabular} &
		\begin{tabular}{@{}c@{}}VST (58 participants) \\ Acc. VT=100 °/s,\\ MFD= 100 ms\end{tabular} &
		\begin{tabular}{@{}c@{}}GOF (153 participants) \\ Acc. VT=22 °/s,\\ MFD= 100 ms\end{tabular} \\
		\midrule
		0 & 14 & 88.08 $\pm$ 0.19\,\%& 85.77 $\pm$ 0.13\,\% & 91.86  $\pm$ 0.22\,\%& 57.53 $\pm$ 0.53\,\%\\
		1& 33 & 93.71 $\pm$ 0.15\,\% & 91.09 $\pm$ 0.16\,\% & 94.41 $\pm$ 0.13\,\%& 76.55 $\pm$ 0.55\,\%\\
		2& 51 & \bf{95.96}$\pm$ 0.09\,\%& 93.23 $\pm$ 0.13\,\% &\bf{94.72} $\pm$ 0.08\,\% &82.67 $\pm$ 0.52\,\%\\
		3& 69 &95.16$\pm$ 0.12\,\% & 93.12 $\pm$ 0.12\,\%& 93.59 $\pm$ 0.12\,\% &\bf{84.48} $\pm$ 0.44\,\%\\
		4& 87 &95.37$\pm$ 0.08\,\%& \bf{93.39} $\pm$ 0.09\,\%& 91.59 $\pm$ 0.20\,\% & 84.08 $\pm$ 0.46\,\%\\
		5& 105 &94.89$\pm$ 0.10\,\%& 93.04 $\pm$ 0.11\,\% & 90.34 $\pm$ 0.19\,\% & 83.20 $\pm$ 0.48\,\%\\
		\bottomrule
	\end{tabular}
	\endgroup
\end{table*}

\begin{figure}
	
\end{figure}

\begin{figure*}
	
	{
		\begin{subfigure}[b]{.49\linewidth}
			\centering
			\includegraphics[width=\linewidth]{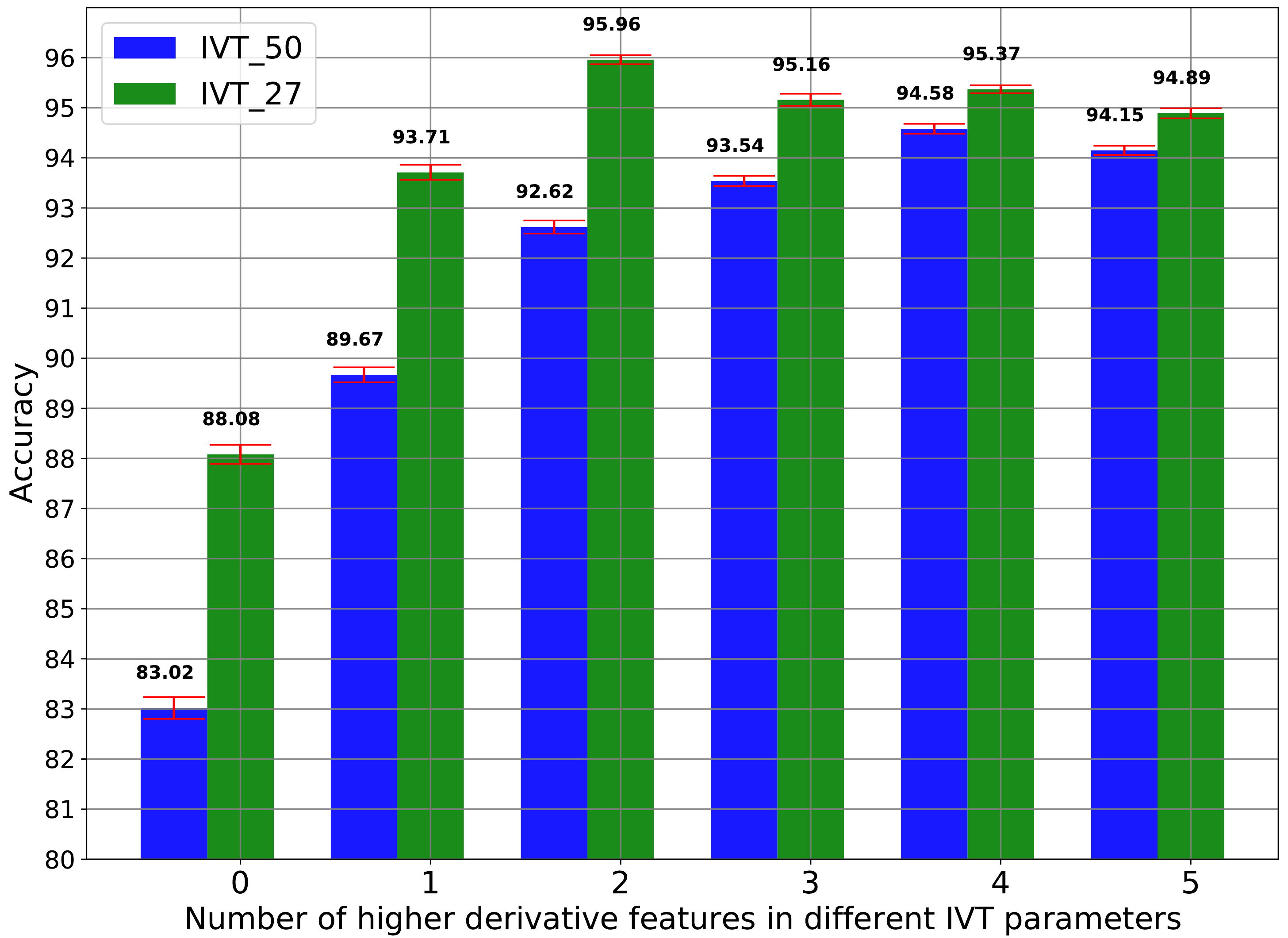}
			\caption{RAN}
			\label{fig_user_prediction_dervatives_features_bioeye_RAN}
		\end{subfigure} 
		\hfill
		\begin{subfigure}[b]{.49\linewidth}
			\centering
			\includegraphics[width=.99\linewidth]{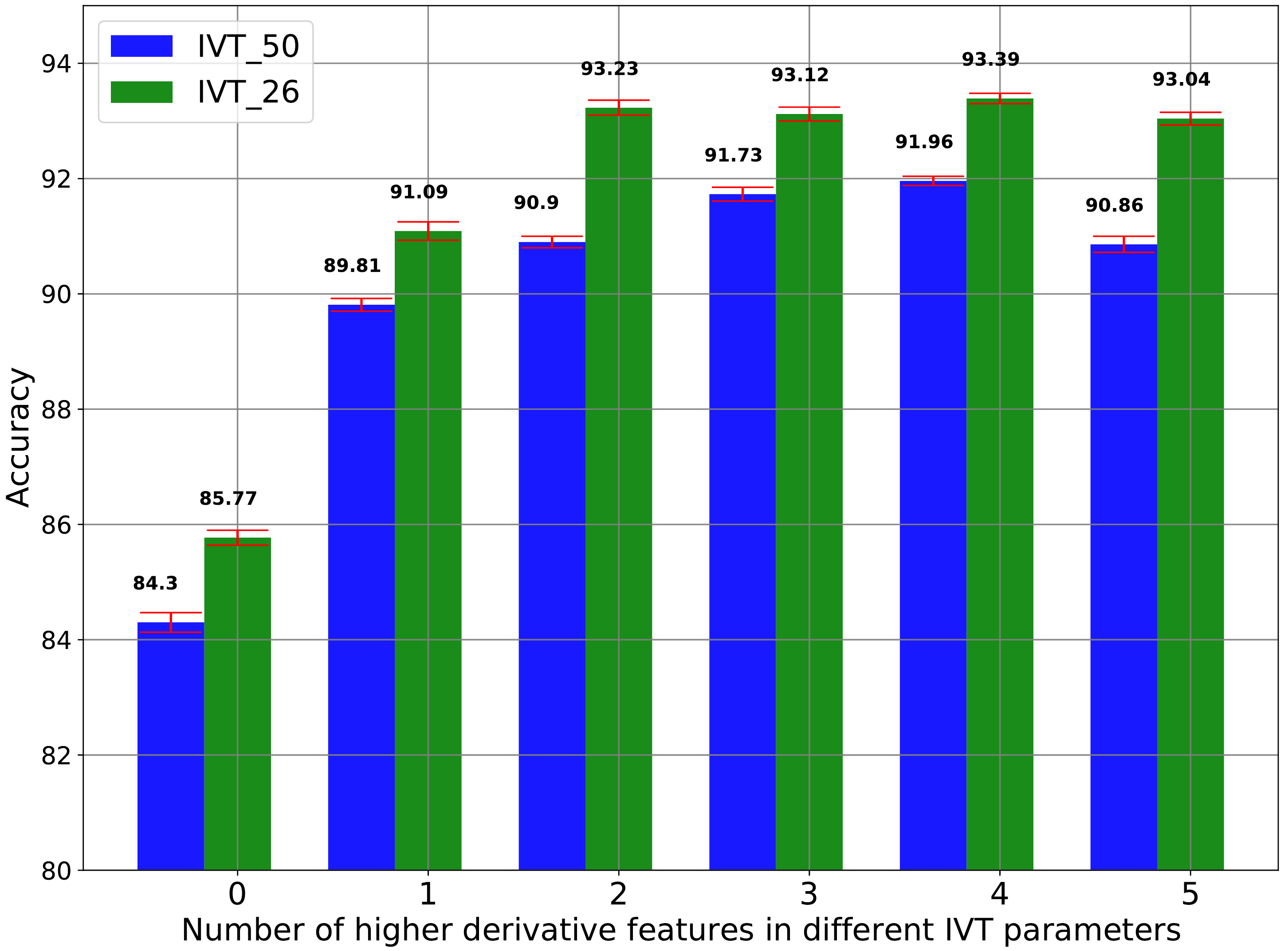}
			\caption{TEX}
			\label{fig_user_prediction_dervatives_features_bioeye_TEX}
		\end{subfigure} 
	}
	
	{
		\begin{subfigure}[b]{.49\linewidth}
			\centering
			\includegraphics[width=.99\linewidth]{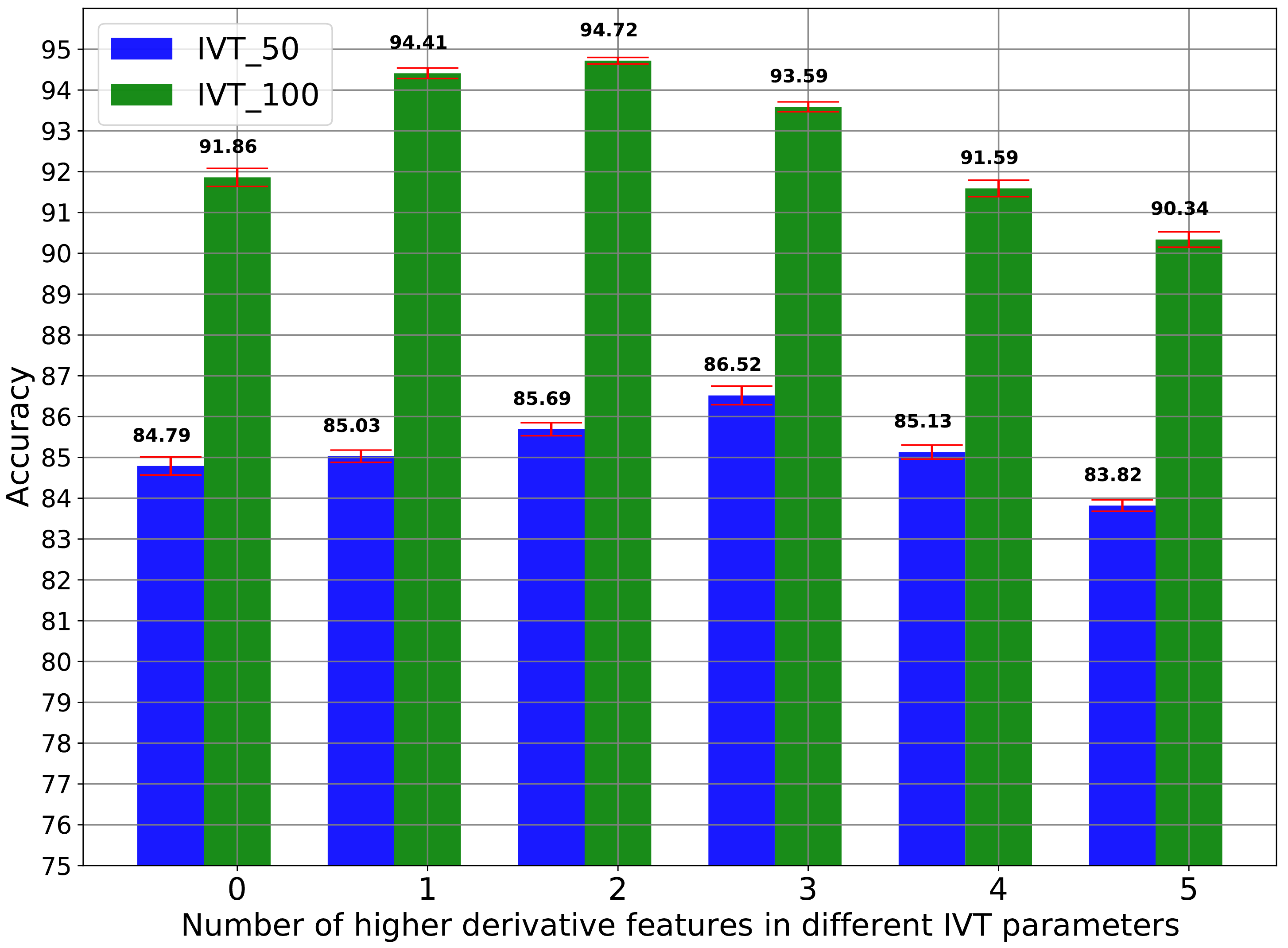}
			\caption{VST}
			\label{fig_user_prediction_dervatives_features_VST}
		\end{subfigure} 
		\hfill
		\begin{subfigure}[b]{.49\linewidth}
			\centering
			\includegraphics[width=.99\linewidth]{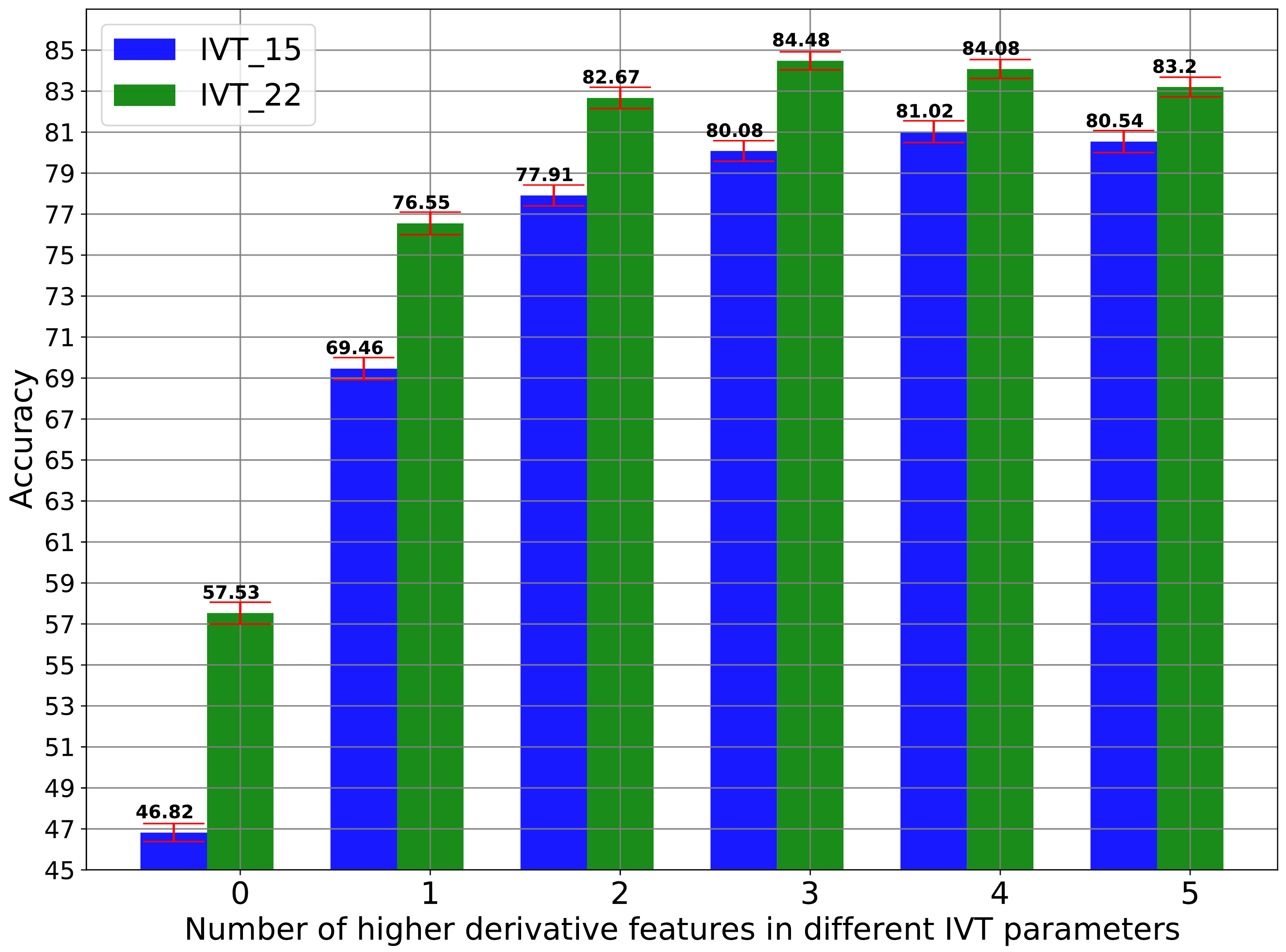}
			\caption{GOF}
			\label{fig_user_prediction_dervatives_features_GOF}
		\end{subfigure} 
	}

	\caption{Performance metrics over 50 runs with different number features of higher order derivatives of the gaze trajectory for different datasets.}
\end{figure*}

\subsection{Combination of IVT Parameter Tuning and Higher-Order Derivatives Features}
\label{sec_combination_ivt_hod}
%
%
Here, we take the best IVT parameters with the approach described in \autoref{IVT_parameters} 
for the four datasets and repeat the user identification experiments with increasing number of higher order derivative features.
\autoref{user_prediction_dervatives_features_all_data_best_VT} shows the average of accuracies over 50 runs of increasing number of higher derivative features of fixation and saccade using the best IVT parameters for the four different datasets.
Similarly, \autoref{fig_user_prediction_dervatives_features_bioeye_RAN}, \subref{fig_user_prediction_dervatives_features_bioeye_TEX}, \subref{fig_user_prediction_dervatives_features_VST}, and \subref{fig_user_prediction_dervatives_features_GOF} compare the accuracies between the default and optimal IVT parameters for RAN, TEX, VST and GOF datasets respectively for including an increasing number of higher order derivative features.
It can be noticed that for the RAN and VST datasets while using the optimal IVT parameters, the best accuracy were observed with fewer features (until acceleration level) in comparison to the case when the default IVT parameters were used (see \autoref{user_prediction_dervatives_features_all_data}).
For TEX and GOF datasets, the best accuracy were achieved with including jounce (93.39 $\pm$ 0.09\,\% for Tex) and jerk (84.48  $\pm$ 0.44\,\% for GOF) level features.

Overall, one can conclude that finding the best IVT parameters and including higher order derivative features are somewhat complementary: When the default threshold is used, including higher order derivatives always increases the best accuracy for all the four datasets.
However, finding the best IVT threshold may yield similar best case accuracies with lower derivative features as this is the case for RAN and VST datasets.
Nevertheless, for TEX and GOF datasets, the accuracy still increases when jounce and jerk level features in TEX and GOF datasets respectively are included even after using the optimal IVT parameters.

\subsection{Combining IVT Tuning and Higher-Order Derivatives with Blink Classifier}
Here, we study the case of including the blink classifier with the optimal IVT threshold and set of higher order derivatives features.
As mentioned in \autoref{Effect_of_blinking_features}, we employ the Nelder-Mead optimization method to find the optimal weights for combining the fixation, saccade and blink classifiers.
The achieved accuracy for the three datasets are reported in \autoref{user_prediction_with_blink_CLF_best_IVT_bioeye_RAN_TEX_VST}.
The accuracy improves by 0.70\,\% for the RAN dataset yielding the final accuracy of 96.64 $\pm$ 0.07\,\%, by 0.15\,\% for the TEX dataset leading to 93.53 $\pm$ 0.11\,\%, and by 0.07\,\% for the VST dataset leading to 94.79 $\pm$ 0.03\,\%.

\begin{table}[ht]
	\centering
	\caption{Performance metrics over 50 runs using blink classifier in RAN data (VT = 27\,°/s and MFD = 96 ms), TEX data (VT = 26\,°/s and MFD = 98 ms), and VST data (VT = 100\,°/s and MFD = 100 ms).}
	\label{user_prediction_with_blink_CLF_best_IVT_bioeye_RAN_TEX_VST}
	\begingroup
	\setlength{\tabcolsep}{2pt} 
	\renewcommand{\arraystretch}{1} 
	\begin{tabular}{cccc}
		\toprule
		 Dataset&\begin{tabular}{@{}c@{}}Fix/Sac/blink\\features \end{tabular} & \begin{tabular}{@{}c@{}}CLF weights\\ Sac/Fix/Blink\end{tabular} & \begin{tabular}{@{}c@{}}Identification\\Accuracy\end{tabular} \\
		\midrule
		RAN & 51/51/7 & 0.5/0.5/0.0 & 95.96 $\pm$ 0.09\,\% \\
		 & 51/51/7 & 0.543/0.447/0.010 & 96.64 $\pm$ 0.07\,\%\\
		\bottomrule
		 TEX & 87/87/0 & 0.5/0.5/0.0 & 93.39 $\pm$ 0.09\,\%\\
		 & 87/87/7 & 0.529/0.466/0.005 & 93.53 $\pm$ 0.11\,\%\\
		\bottomrule
		VST & 51/51/0 & 0.5/0.5/0.0 & 94.72 $\pm$ 0.08\,\%\\
		& 51/51/7 & 0.513/0.477/0.010 & 94.79 $\pm$ 0.03\,\%\\
		\bottomrule
	\end{tabular}
	\endgroup
\end{table}

\begin{table}
	\centering
	\caption{A comparison of performance metrics over 50 runs using 51 features with 58 participants of all the datasets.}
	\label{compare_user_prediction_in_all_data}
	\begingroup
	\setlength{\tabcolsep}{6pt} 
	\renewcommand{\arraystretch}{1.1} 
	\begin{tabular}{ccc}
		\toprule
		\begin{tabular}{@{}c@{}}Datasets\end{tabular} & \begin{tabular}{@{}c@{}}VT, (MFD = 100\,ms)\end{tabular} &\specialcell{Identification\\Accuracy}\\
		\midrule
		RAN & $50\,^{\circ}\!/s$ &90.37 $\pm$ 0.48\,\%\\
		TEX & $50\,^{\circ}\!/s$ &90.90 $\pm$ 0.10\,\%\\
		VST & $50\,^{\circ}\!/s$ &73.59 $\pm$ 0.39\,\%\\
		GOF & $15\,^{\circ}\!/s$ & 84.48 $\pm$ 0.79\,\%\\
		\bottomrule
		RAN & $27\,^{\circ}\!/s$ &92.82 $\pm$ 0.38\,\%\\
		TEX & $26\,^{\circ}\!/s$ &93.23 $\pm$ 0.13\,\%\\
		VST & $100\,^{\circ}\!/s$&71.62 $\pm$ 0.35\,\%\\
		GOF & $22\,^{\circ}\!/s$ &85.21 $\pm$ 0.64\,\%\\
		\bottomrule
	\end{tabular}
	\endgroup
\end{table}

\begin{table}
	\caption{Identification accuracy, number of fixations in BioEye data sets with two different VT and different trajectory length of RAN data.}
	\label{table_same_trajectory_length}
	\begingroup
	\setlength{\tabcolsep}{3.5pt} 
	\renewcommand{\arraystretch}{1} 
	\begin{tabular}{cccc}
		\toprule
		& RAN & (MFD = 100\,ms)& \\
		\midrule
		Vel. & Fix. No. & \specialcell{Identification\\Accuracy}  & Trajectory \\
		threshold & & &length \\
		\midrule
		50 &16891&84.43 $\pm$ 0.23\,\%& 60 sec (start) \\	
		27 &18831&89.75 $\pm$ 0.16\,\%& 60 sec (start)\\		
		50 &17451&82.37 $\pm$ 0.20\,\%& 60 sec (end)\\
		\bottomrule
		& TEX & (MFD = 100 ms)&  \\
		\bottomrule
		50 &32944&90.90 $\pm$ 0.10\,\%& 60 sec\\
		26 &33584&93.23 $\pm$ 0.13\,\%& 60 sec\\
		\bottomrule
	\end{tabular}
	\endgroup
\end{table}

\subsection{Effect of Stimuli after Homogenizing the Different Datasets}
\label{trajectory_len_stimuli}
Here we revisit the question which stimuli is the best for user prediction.
This was previously studied in \autoref{sec_effect_of_stimuli}.
To get a fair comparison we homogenize the different aspects, i.e., number of participants, trajectory length, and age group of these datasets.
As mentioned previously, each dataset has a different number of participants (RAN: 153, TEX: 153, VST: 58 and GOF: 378).
For a fair comparison, the minimum number of participants among all the datasets has been selected (i.e. 58 participants).
The participants are drawn at random when the datasets have a higher number of available participants.
A fixed trajectory length (first minute) is selected to test the effect of the stimuli on the user identification accuracy.
Since, RAN, TEX and VST datasets have participants only in the age group 20--40, we selected only participants of the corresponding age group 1 of the GOF dataset.

The results of these experiments are reported in~\autoref{compare_user_prediction_in_all_data}.
One can notice that the TEX data accuracy is the highest: 93.23\,\%  with the optimal IVT parameter VT = 26\,°/s.
This is according to the state of art e.g.~\cite{friedman2017method,lohr2020eye} since the eye movements during reading have been used to achieve some of the best biometric performances.
However, it is important to consider the number of fixations (if recorded with the same eye tracker) before concluding, which stimuli is the best, as the number of fixations plays a vital role in attaining a higher accuracy.
The higher the number of fixations, the better the accuracy achieved.
~\autoref{table_same_trajectory_length} notes the number of fixations for the RAN datasets for a trajectory length of 1 minute.
It is much lower than in the TEX dataset, which might influence that the accuracy for TEX is higher than that of RAN.

%% file: content/Conclusion.tex
\section{Conclusion and future work}
\label{Conclusion}

This paper presents an approach for user identification which works consistently and robustly well with four different datasets of varying stimuli.
The achieved identification accuracies are 96.64\,\% in RAN, 93.53\,\% in TEX, 94.72\,\% in VST, and 84.48\,\% in GOF data.
Additionally, an extensive study has been done to investigate different factors (for example, IVT parameters, higher order derivative features, effect of gender and age, template aging effect, blinks etc.) that can affect user identification performance.
Our results suggest that selecting the best VT and higher-order derivative features have the greatest impact on accuracy.
We also found that user identification works better in the solo female group than in solo male groups and hence is biased towards gender.
Similarly, it works better in the older age group than in a younger age group of participants.
We believe more work is needed on improving the accuracy of user identification when there is a significant time gap between train and test sessions.


%% file: content/Acknowledgment.tex
\section*{Acknowledgment}
The first author would like to acknowledge the support of DAAD PhD scholarship (award number 91645228) for funding this research.

%% file: main.bbl

\begin{thebibliography}{44}


\ifx \showCODEN    \undefined \def \showCODEN     #1{\unskip}     \fi
\ifx \showDOI      \undefined \def \showDOI       #1{#1}\fi
\ifx \showISBNx    \undefined \def \showISBNx     #1{\unskip}     \fi
\ifx \showISBNxiii \undefined \def \showISBNxiii  #1{\unskip}     \fi
\ifx \showISSN     \undefined \def \showISSN      #1{\unskip}     \fi
\ifx \showLCCN     \undefined \def \showLCCN      #1{\unskip}     \fi
\ifx \shownote     \undefined \def \shownote      #1{#1}          \fi
\ifx \showarticletitle \undefined \def \showarticletitle #1{#1}   \fi
\ifx \showURL      \undefined \def \showURL       {\relax}        \fi
\providecommand\bibfield[2]{#2}
\providecommand\bibinfo[2]{#2}
\providecommand\natexlab[1]{#1}
\providecommand\showeprint[2][]{arXiv:#2}

\bibitem[\protect\citeauthoryear{Alkan and Cagiltay}{Alkan and
  Cagiltay}{2007}]%
        {alkan2007studying}
\bibfield{author}{\bibinfo{person}{Serkan Alkan} {and} \bibinfo{person}{Kursat
  Cagiltay}.} \bibinfo{year}{2007}\natexlab{}.
\newblock \showarticletitle{Studying computer game learning experience through
  eye tracking}.
\newblock \bibinfo{journal}{\emph{British Journal of Educational Technology}}
  \bibinfo{volume}{38}, \bibinfo{number}{3} (\bibinfo{year}{2007}),
  \bibinfo{pages}{538--542}.
\newblock


\bibitem[\protect\citeauthoryear{Andersson, Larsson, Holmqvist, Stridh, and
  Nystr{\"o}m}{Andersson et~al\mbox{.}}{2017}]%
        {Andersson2017}
\bibfield{author}{\bibinfo{person}{Richard Andersson}, \bibinfo{person}{Linnea
  Larsson}, \bibinfo{person}{Kenneth Holmqvist}, \bibinfo{person}{Martin
  Stridh}, {and} \bibinfo{person}{Marcus Nystr{\"o}m}.}
  \bibinfo{year}{2017}\natexlab{}.
\newblock \showarticletitle{One algorithm to rule them all? An evaluation and
  discussion of ten eye movement event-detection algorithms}.
\newblock \bibinfo{journal}{\emph{Behavior Research Methods}}
  \bibinfo{volume}{49}, \bibinfo{number}{2} (\bibinfo{year}{2017}),
  \bibinfo{pages}{616--637}.
\newblock
\showISSN{1554-3528}
\urldef\tempurl%
\url{https://doi.org/10.3758/s13428-016-0738-9}
\showDOI{\tempurl}


\bibitem[\protect\citeauthoryear{Armstrong and Olatunji}{Armstrong and
  Olatunji}{2012}]%
        {armstrong2012eye}
\bibfield{author}{\bibinfo{person}{Thomas Armstrong} {and}
  \bibinfo{person}{Bunmi~O. Olatunji}.} \bibinfo{year}{2012}\natexlab{}.
\newblock \showarticletitle{Eye tracking of attention in the affective
  disorders: A meta-analytic review and synthesis}.
\newblock \bibinfo{journal}{\emph{Clinical psychology review}}
  \bibinfo{volume}{32}, \bibinfo{number}{8} (\bibinfo{year}{2012}),
  \bibinfo{pages}{704--723}.
\newblock


\bibitem[\protect\citeauthoryear{Benfatto, Seimyr, Ygge, Pansell, Rydberg, and
  Jacobson}{Benfatto et~al\mbox{.}}{2016}]%
        {benfatto2016screening}
\bibfield{author}{\bibinfo{person}{Mattias~Nilsson Benfatto},
  \bibinfo{person}{Gustaf~{\"O}qvist Seimyr}, \bibinfo{person}{Jan Ygge},
  \bibinfo{person}{Tony Pansell}, \bibinfo{person}{Agneta Rydberg}, {and}
  \bibinfo{person}{Christer Jacobson}.} \bibinfo{year}{2016}\natexlab{}.
\newblock \showarticletitle{Screening for dyslexia using eye tracking during
  reading}.
\newblock \bibinfo{journal}{\emph{PloS one}} \bibinfo{volume}{11},
  \bibinfo{number}{12} (\bibinfo{year}{2016}), \bibinfo{pages}{e0165508}.
\newblock


\bibitem[\protect\citeauthoryear{Billeci, Narzisi, Tonacci, Sbriscia-Fioretti,
  Serasini, Fulceri, Apicella, Sicca, Calderoni, and Muratori}{Billeci
  et~al\mbox{.}}{2017}]%
        {billeci2017integrated}
\bibfield{author}{\bibinfo{person}{Lucia Billeci}, \bibinfo{person}{Antonio
  Narzisi}, \bibinfo{person}{Alessandro Tonacci}, \bibinfo{person}{Beatrice
  Sbriscia-Fioretti}, \bibinfo{person}{Luca Serasini},
  \bibinfo{person}{Francesca Fulceri}, \bibinfo{person}{Fabio Apicella},
  \bibinfo{person}{Federico Sicca}, \bibinfo{person}{Sara Calderoni}, {and}
  \bibinfo{person}{Filippo Muratori}.} \bibinfo{year}{2017}\natexlab{}.
\newblock \showarticletitle{An integrated EEG and eye-tracking approach for the
  study of responding and initiating joint attention in Autism Spectrum
  Disorders}.
\newblock \bibinfo{journal}{\emph{Scientific Reports}} \bibinfo{volume}{7},
  \bibinfo{number}{1} (\bibinfo{year}{2017}), \bibinfo{pages}{1--13}.
\newblock


\bibitem[\protect\citeauthoryear{Breiman}{Breiman}{2001}]%
        {DBLP:journals/ml/Breiman01}
\bibfield{author}{\bibinfo{person}{Leo Breiman}.}
  \bibinfo{year}{2001}\natexlab{}.
\newblock \showarticletitle{Random Forests}.
\newblock \bibinfo{journal}{\emph{Machine Learning}} \bibinfo{volume}{45},
  \bibinfo{number}{1} (\bibinfo{year}{2001}), \bibinfo{pages}{5--32}.
\newblock
\urldef\tempurl%
\url{https://doi.org/10.1023/A:1010933404324}
\showDOI{\tempurl}


\bibitem[\protect\citeauthoryear{Broomhead and Lowe}{Broomhead and
  Lowe}{1988}]%
        {broomhead1988radial}
\bibfield{author}{\bibinfo{person}{David~S. Broomhead} {and}
  \bibinfo{person}{David Lowe}.} \bibinfo{year}{1988}\natexlab{}.
\newblock \bibinfo{booktitle}{\emph{Radial basis functions, multi-variable
  functional interpolation and adaptive networks}}.
\newblock \bibinfo{type}{{T}echnical {R}eport}. \bibinfo{institution}{Royal
  Signals and Radar Establishment Malvern (United Kingdom)}.
\newblock


\bibitem[\protect\citeauthoryear{Coutrot, Binetti, Harrison, Mareschal, and
  Johnston}{Coutrot et~al\mbox{.}}{2016}]%
        {coutrot2016face}
\bibfield{author}{\bibinfo{person}{Antoine Coutrot}, \bibinfo{person}{Nicola
  Binetti}, \bibinfo{person}{Charlotte Harrison}, \bibinfo{person}{Isabelle
  Mareschal}, {and} \bibinfo{person}{Alan Johnston}.}
  \bibinfo{year}{2016}\natexlab{}.
\newblock \showarticletitle{Face exploration dynamics differentiate men and
  women}.
\newblock \bibinfo{journal}{\emph{Journal of vision}} \bibinfo{volume}{16},
  \bibinfo{number}{14} (\bibinfo{year}{2016}), \bibinfo{pages}{16--16}.
\newblock


\bibitem[\protect\citeauthoryear{Doughty}{Doughty}{2002}]%
        {doughty2002further}
\bibfield{author}{\bibinfo{person}{Michael~J. Doughty}.}
  \bibinfo{year}{2002}\natexlab{}.
\newblock \showarticletitle{Further assessment of gender-and blink
  pattern-related differences in the spontaneous eyeblink activity in primary
  gaze in young adult humans}.
\newblock \bibinfo{journal}{\emph{Optometry and Vision Science}}
  \bibinfo{volume}{79}, \bibinfo{number}{7} (\bibinfo{year}{2002}),
  \bibinfo{pages}{439--447}.
\newblock


\bibitem[\protect\citeauthoryear{Esfahani}{Esfahani}{2016}]%
        {Esf16}
\bibfield{author}{\bibinfo{person}{Nastaran~Maus Esfahani}.}
  \bibinfo{year}{2016}\natexlab{}.
\newblock \showarticletitle{A Brief Review of Human Identification Using Eye
  Movement}.
\newblock \bibinfo{journal}{\emph{Journal of Pattern Recognition Research}}
  \bibinfo{volume}{11}, \bibinfo{number}{1} (\bibinfo{year}{2016}),
  \bibinfo{pages}{15--24}.
\newblock
\urldef\tempurl%
\url{https://doi.org/10.13140/RG.2.1.3466.3924}
\showDOI{\tempurl}


\bibitem[\protect\citeauthoryear{Friedman, Nixon, and Komogortsev}{Friedman
  et~al\mbox{.}}{2017}]%
        {friedman2017method}
\bibfield{author}{\bibinfo{person}{Lee Friedman}, \bibinfo{person}{Mark~S.
  Nixon}, {and} \bibinfo{person}{Oleg~V. Komogortsev}.}
  \bibinfo{year}{2017}\natexlab{}.
\newblock \showarticletitle{Method to assess the temporal persistence of
  potential biometric features: Application to oculomotor, gait, face and brain
  structure databases}.
\newblock \bibinfo{journal}{\emph{PloS one}} \bibinfo{volume}{12},
  \bibinfo{number}{6} (\bibinfo{year}{2017}), \bibinfo{pages}{e0178501}.
\newblock


\bibitem[\protect\citeauthoryear{Galdi, Nappi, Riccio, and Wechsler}{Galdi
  et~al\mbox{.}}{2016}]%
        {DBLP:journals/prl/GaldiNRW16}
\bibfield{author}{\bibinfo{person}{Chiara Galdi}, \bibinfo{person}{Michele
  Nappi}, \bibinfo{person}{Daniel Riccio}, {and} \bibinfo{person}{Harry
  Wechsler}.} \bibinfo{year}{2016}\natexlab{}.
\newblock \showarticletitle{Eye movement analysis for human authentication: a
  critical survey}.
\newblock \bibinfo{journal}{\emph{Pattern Recognition Letters}}
  \bibinfo{volume}{84} (\bibinfo{year}{2016}), \bibinfo{pages}{272--283}.
\newblock
\urldef\tempurl%
\url{https://doi.org/10.1016/j.patrec.2016.11.002}
\showDOI{\tempurl}


\bibitem[\protect\citeauthoryear{Gao and Han}{Gao and Han}{2012}]%
        {gao2012implementing}
\bibfield{author}{\bibinfo{person}{Fuchang Gao} {and} \bibinfo{person}{Lixing
  Han}.} \bibinfo{year}{2012}\natexlab{}.
\newblock \showarticletitle{Implementing the Nelder-Mead simplex algorithm with
  adaptive parameters}.
\newblock \bibinfo{journal}{\emph{Computational Optimization and Applications}}
  \bibinfo{volume}{51}, \bibinfo{number}{1} (\bibinfo{year}{2012}),
  \bibinfo{pages}{259--277}.
\newblock


\bibitem[\protect\citeauthoryear{George and Routray}{George and
  Routray}{2016}]%
        {george2016score}
\bibfield{author}{\bibinfo{person}{Anjith George} {and}
  \bibinfo{person}{Aurobinda Routray}.} \bibinfo{year}{2016}\natexlab{}.
\newblock \showarticletitle{A score level fusion method for eye movement
  biometrics}.
\newblock \bibinfo{journal}{\emph{Pattern Recognition Letters}}
  \bibinfo{volume}{82} (\bibinfo{year}{2016}), \bibinfo{pages}{207--215}.
\newblock
\urldef\tempurl%
\url{https://doi.org/10.1016/j.patrec.2015.11.020}
\showDOI{\tempurl}


\bibitem[\protect\citeauthoryear{Holland and Komogortsev}{Holland and
  Komogortsev}{2011}]%
        {DBLP:conf/icb/HollandK11}
\bibfield{author}{\bibinfo{person}{Corey Holland} {and}
  \bibinfo{person}{Oleg~V. Komogortsev}.} \bibinfo{year}{2011}\natexlab{}.
\newblock \showarticletitle{Biometric identification via eye movement scanpaths
  in reading}. In \bibinfo{booktitle}{\emph{2011 {IEEE} International Joint
  Conference on Biometrics, {IJCB} 2011, Washington, DC, USA, October 11-13,
  2011}}. \bibinfo{pages}{1--8}.
\newblock
\urldef\tempurl%
\url{https://doi.org/10.1109/IJCB.2011.6117536}
\showDOI{\tempurl}


\bibitem[\protect\citeauthoryear{Holland and Komogortsev}{Holland and
  Komogortsev}{2012}]%
        {DBLP:conf/btas/HollandK12}
\bibfield{author}{\bibinfo{person}{Corey~D. Holland} {and}
  \bibinfo{person}{Oleg~V. Komogortsev}.} \bibinfo{year}{2012}\natexlab{}.
\newblock \showarticletitle{Biometric verification via complex eye movements:
  The effects of environment and stimulus}. In \bibinfo{booktitle}{\emph{{IEEE}
  Fifth International Conference on Biometrics: Theory, Applications and
  Systems, {BTAS} 2012, Arlington, VA, USA, September 23-27, 2012}}.
  \bibinfo{pages}{39--46}.
\newblock
\urldef\tempurl%
\url{https://doi.org/10.1109/BTAS.2012.6374556}
\showDOI{\tempurl}


\bibitem[\protect\citeauthoryear{J{\"a}ger, Makowski, Prasse, Liehr, Seidler,
  and Scheffer}{J{\"a}ger et~al\mbox{.}}{2019}]%
        {jager2019deep}
\bibfield{author}{\bibinfo{person}{Lena~A. J{\"a}ger}, \bibinfo{person}{Silvia
  Makowski}, \bibinfo{person}{Paul Prasse}, \bibinfo{person}{Sascha Liehr},
  \bibinfo{person}{Maximilian Seidler}, {and} \bibinfo{person}{Tobias
  Scheffer}.} \bibinfo{year}{2019}\natexlab{}.
\newblock \showarticletitle{Deep Eyedentification: Biometric identification
  using micro-movements of the eye}. In \bibinfo{booktitle}{\emph{Joint
  European Conference on Machine Learning and Knowledge Discovery in
  Databases}}. Springer, \bibinfo{pages}{299--314}.
\newblock


\bibitem[\protect\citeauthoryear{Juan}{Juan}{2006}]%
        {blinkINinfantLess}
\bibfield{author}{\bibinfo{person}{Stephen Juan}.} \bibinfo{year}{30 Jun
  2006}\natexlab{}.
\newblock \bibinfo{title}{{Why do babies blink less often than adults?}}
\newblock
  \bibinfo{howpublished}{\url{https://www.theregister.com/2006/06/30/the_odd_body_blinking/}}.
\newblock
\newblock
\shownote{[Online; accessed 11-August-2021].}


\bibitem[\protect\citeauthoryear{Kasneci, Kasneci, Trautwein, Appel, Tibus,
  Jaeggi, and Gerjets}{Kasneci et~al\mbox{.}}{2021}]%
        {kasneci2021your}
\bibfield{author}{\bibinfo{person}{Enkelejda Kasneci}, \bibinfo{person}{Gjergji
  Kasneci}, \bibinfo{person}{Ulrich Trautwein}, \bibinfo{person}{Tobias Appel},
  \bibinfo{person}{Maike Tibus}, \bibinfo{person}{Susanne~M. Jaeggi}, {and}
  \bibinfo{person}{Peter Gerjets}.} \bibinfo{year}{2021}\natexlab{}.
\newblock \showarticletitle{Do your eye movements reveal your performance on an
  IQ test? A study linking eye movements and socio-demographic information to
  fluid intelligence.}
\newblock \bibinfo{journal}{\emph{PsyArXiv}} (\bibinfo{date}{Mar}
  \bibinfo{year}{2021}).
\newblock
\urldef\tempurl%
\url{https://doi.org/10.31234/osf.io/dru93}
\showDOI{\tempurl}


\bibitem[\protect\citeauthoryear{Kasprowski and Ober}{Kasprowski and
  Ober}{2004}]%
        {DBLP:conf/eccv/KasprowskiO04}
\bibfield{author}{\bibinfo{person}{Pawel Kasprowski} {and}
  \bibinfo{person}{J{\'{o}}zef Ober}.} \bibinfo{year}{2004}\natexlab{}.
\newblock \showarticletitle{Eye Movements in Biometrics}. In
  \bibinfo{booktitle}{\emph{Biometric Authentication, {ECCV} 2004 International
  Workshop, BioAW 2004, Prague, Czech Republic, May 15, 2004, Proceedings}}.
  \bibinfo{pages}{248--258}.
\newblock
\urldef\tempurl%
\url{https://doi.org/10.1007/978-3-540-25976-3\_23}
\showDOI{\tempurl}


\bibitem[\protect\citeauthoryear{Krishna, Ding, Xu, and H{\"o}llerer}{Krishna
  et~al\mbox{.}}{2019}]%
        {krishna2019multimodal}
\bibfield{author}{\bibinfo{person}{Vrishab Krishna}, \bibinfo{person}{Yi Ding},
  \bibinfo{person}{Aiwen Xu}, {and} \bibinfo{person}{Tobias H{\"o}llerer}.}
  \bibinfo{year}{2019}\natexlab{}.
\newblock \showarticletitle{Multimodal biometric authentication for VR/AR using
  EEG and eye tracking}. In \bibinfo{booktitle}{\emph{Adjunct of the 2019
  International Conference on Multimodal Interaction}}. \bibinfo{pages}{1--5}.
\newblock


\bibitem[\protect\citeauthoryear{Kr{\"o}ger, Lutz, and M{\"u}ller}{Kr{\"o}ger
  et~al\mbox{.}}{2019}]%
        {kroger2019does}
\bibfield{author}{\bibinfo{person}{Jacob~Leon Kr{\"o}ger},
  \bibinfo{person}{Otto Hans-Martin Lutz}, {and} \bibinfo{person}{Florian
  M{\"u}ller}.} \bibinfo{year}{2019}\natexlab{}.
\newblock \showarticletitle{What does your gaze reveal about you? On the
  privacy implications of eye tracking}. In \bibinfo{booktitle}{\emph{IFIP
  International Summer School on Privacy and Identity Management}}. Springer,
  \bibinfo{pages}{226--241}.
\newblock


\bibitem[\protect\citeauthoryear{Lankes and Stoeckl}{Lankes and
  Stoeckl}{2020}]%
        {lankes2020gazing}
\bibfield{author}{\bibinfo{person}{Michael Lankes} {and}
  \bibinfo{person}{Andreas Stoeckl}.} \bibinfo{year}{2020}\natexlab{}.
\newblock \showarticletitle{Gazing at Pac-Man: Lessons Learned from a
  Eye-Tracking Study Focusing on Game Difficulty}. In
  \bibinfo{booktitle}{\emph{ACM Symposium on Eye Tracking Research and
  Applications}} (Stuttgart, Germany) \emph{(\bibinfo{series}{ETRA '20 Short
  Papers})}. \bibinfo{publisher}{Association for Computing Machinery},
  \bibinfo{address}{New York, NY, USA}, Article \bibinfo{articleno}{62},
  \bibinfo{numpages}{5}~pages.
\newblock
\showISBNx{9781450371346}
\urldef\tempurl%
\url{https://doi.org/10.1145/3379156.3391840}
\showDOI{\tempurl}


\bibitem[\protect\citeauthoryear{Li, Xue, Quan, Yue, and Zhang}{Li
  et~al\mbox{.}}{2018}]%
        {li2018biometrictask}
\bibfield{author}{\bibinfo{person}{Chunyong Li}, \bibinfo{person}{Jiguo Xue},
  \bibinfo{person}{Cheng Quan}, \bibinfo{person}{Jingwei Yue}, {and}
  \bibinfo{person}{Chenggang Zhang}.} \bibinfo{year}{2018}\natexlab{}.
\newblock \showarticletitle{Biometric recognition via texture features of eye
  movement trajectories in a visual searching task}.
\newblock \bibinfo{journal}{\emph{PloS one}} \bibinfo{volume}{13},
  \bibinfo{number}{4} (\bibinfo{year}{2018}), \bibinfo{pages}{e0194475}.
\newblock


\bibitem[\protect\citeauthoryear{Lin, Huan, Chan, Yeh, and Chiu}{Lin
  et~al\mbox{.}}{2004}]%
        {lin2004design}
\bibfield{author}{\bibinfo{person}{Chern-Sheng Lin}, \bibinfo{person}{Chia-Chin
  Huan}, \bibinfo{person}{Chao-Ning Chan}, \bibinfo{person}{Mau-Shiun Yeh},
  {and} \bibinfo{person}{Chuang-Chien Chiu}.} \bibinfo{year}{2004}\natexlab{}.
\newblock \showarticletitle{Design of a computer game using an eye-tracking
  device for eye's activity rehabilitation}.
\newblock \bibinfo{journal}{\emph{Optics and lasers in engineering}}
  \bibinfo{volume}{42}, \bibinfo{number}{1} (\bibinfo{year}{2004}),
  \bibinfo{pages}{91--108}.
\newblock


\bibitem[\protect\citeauthoryear{Lohr, Aziz, and Komogortsev}{Lohr
  et~al\mbox{.}}{2020}]%
        {lohr2020eye}
\bibfield{author}{\bibinfo{person}{Dillon~J. Lohr}, \bibinfo{person}{Samantha
  Aziz}, {and} \bibinfo{person}{Oleg Komogortsev}.}
  \bibinfo{year}{2020}\natexlab{}.
\newblock \showarticletitle{Eye Movement Biometrics Using a New Dataset
  Collected in Virtual Reality}. In \bibinfo{booktitle}{\emph{ACM Symposium on
  Eye Tracking Research and Applications}} (Stuttgart, Germany)
  \emph{(\bibinfo{series}{ETRA '20 Adjunct})}. \bibinfo{publisher}{Association
  for Computing Machinery}, \bibinfo{address}{New York, NY, USA}, Article
  \bibinfo{articleno}{40}, \bibinfo{numpages}{3}~pages.
\newblock
\showISBNx{9781450371353}
\urldef\tempurl%
\url{https://doi.org/10.1145/3379157.3391420}
\showDOI{\tempurl}


\bibitem[\protect\citeauthoryear{Moss, Baddeley, and Canagarajah}{Moss
  et~al\mbox{.}}{2012}]%
        {moss2012eye}
\bibfield{author}{\bibinfo{person}{Felix Joseph~Mercer Moss},
  \bibinfo{person}{Roland Baddeley}, {and} \bibinfo{person}{Nishan
  Canagarajah}.} \bibinfo{year}{2012}\natexlab{}.
\newblock \showarticletitle{Eye movements to natural images as a function of
  sex and personality}.
\newblock \bibinfo{journal}{\emph{PLoS One}} \bibinfo{volume}{7},
  \bibinfo{number}{11} (\bibinfo{year}{2012}), \bibinfo{pages}{e47870}.
\newblock


\bibitem[\protect\citeauthoryear{Nelder and Mead}{Nelder and Mead}{1965}]%
        {1965_nelder_mead}
\bibfield{author}{\bibinfo{person}{J.~A. Nelder} {and} \bibinfo{person}{R.
  Mead}.} \bibinfo{year}{1965}\natexlab{}.
\newblock \showarticletitle{{A Simplex Method for Function Minimization}}.
\newblock \bibinfo{journal}{\emph{Comput. J.}} \bibinfo{volume}{7},
  \bibinfo{number}{4} (\bibinfo{date}{01} \bibinfo{year}{1965}),
  \bibinfo{pages}{308--313}.
\newblock
\showISSN{0010-4620}
\urldef\tempurl%
\url{https://doi.org/10.1093/comjnl/7.4.308}
\showDOI{\tempurl}
\showeprint{https://academic.oup.com/comjnl/article-pdf/7/4/308/1013182/7-4-308.pdf}


\bibitem[\protect\citeauthoryear{Olsen}{Olsen}{2012}]%
        {olsen2012tobii}
\bibfield{author}{\bibinfo{person}{Anneli Olsen}.}
  \bibinfo{year}{2012}\natexlab{}.
\newblock \showarticletitle{The Tobii I-VT fixation filter}.
\newblock \bibinfo{journal}{\emph{Tobii Technology}}  \bibinfo{volume}{21}
  (\bibinfo{year}{2012}).
\newblock


\bibitem[\protect\citeauthoryear{Olsen and Matos}{Olsen and Matos}{2012}]%
        {DBLP:conf/etra/OlsenM12}
\bibfield{author}{\bibinfo{person}{Anneli Olsen} {and} \bibinfo{person}{Ricardo
  Matos}.} \bibinfo{year}{2012}\natexlab{}.
\newblock \showarticletitle{Identifying parameter values for an {I-VT} fixation
  filter suitable for handling data sampled with various sampling frequencies}.
  In \bibinfo{booktitle}{\emph{Proceedings of the 2012 Symposium on
  Eye-Tracking Research and Applications, {ETRA} 2012, Santa Barbara, CA, USA,
  March 28-30, 2012}}. \bibinfo{pages}{317--320}.
\newblock
\urldef\tempurl%
\url{https://doi.org/10.1145/2168556.2168625}
\showDOI{\tempurl}


\bibitem[\protect\citeauthoryear{Pedregosa, Varoquaux, Gramfort, Michel,
  Thirion, Grisel, Blondel, Prettenhofer, Weiss, Dubourg, Vanderplas, Passos,
  Cournapeau, Brucher, Perrot, and Duchesnay}{Pedregosa et~al\mbox{.}}{2011}]%
        {scikit-learn}
\bibfield{author}{\bibinfo{person}{F. Pedregosa}, \bibinfo{person}{G.
  Varoquaux}, \bibinfo{person}{A. Gramfort}, \bibinfo{person}{V. Michel},
  \bibinfo{person}{B. Thirion}, \bibinfo{person}{O. Grisel},
  \bibinfo{person}{M. Blondel}, \bibinfo{person}{P. Prettenhofer},
  \bibinfo{person}{R. Weiss}, \bibinfo{person}{V. Dubourg}, \bibinfo{person}{J.
  Vanderplas}, \bibinfo{person}{A. Passos}, \bibinfo{person}{D. Cournapeau},
  \bibinfo{person}{M. Brucher}, \bibinfo{person}{M. Perrot}, {and}
  \bibinfo{person}{E. Duchesnay}.} \bibinfo{year}{2011}\natexlab{}.
\newblock \showarticletitle{Scikit-learn: Machine Learning in {P}ython}.
\newblock \bibinfo{journal}{\emph{Journal of Machine Learning Research}}
  \bibinfo{volume}{12} (\bibinfo{year}{2011}), \bibinfo{pages}{2825--2830}.
\newblock


\bibitem[\protect\citeauthoryear{Rakoczi, Duchowski, Casas-Tost, and
  Pohl}{Rakoczi et~al\mbox{.}}{2013}]%
        {rakoczi2013visual}
\bibfield{author}{\bibinfo{person}{Gergely Rakoczi}, \bibinfo{person}{Andrew
  Duchowski}, \bibinfo{person}{Helena Casas-Tost}, {and}
  \bibinfo{person}{Margit Pohl}.} \bibinfo{year}{2013}\natexlab{}.
\newblock \showarticletitle{Visual perception of international traffic signs:
  influence of e-learning and culture on eye movements}. In
  \bibinfo{booktitle}{\emph{Proceedings of the 2013 Conference on Eye Tracking
  South Africa}}. \bibinfo{pages}{8--16}.
\newblock


\bibitem[\protect\citeauthoryear{Rigas and Komogortsev}{Rigas and
  Komogortsev}{2017}]%
        {DBLP:journals/ivc/RigasK17}
\bibfield{author}{\bibinfo{person}{Ioannis Rigas} {and}
  \bibinfo{person}{Oleg~V. Komogortsev}.} \bibinfo{year}{2017}\natexlab{}.
\newblock \showarticletitle{Current research in eye movement biometrics: An
  analysis based on BioEye 2015 competition}.
\newblock \bibinfo{journal}{\emph{Image Vision Computing}}
  \bibinfo{volume}{58} (\bibinfo{year}{2017}), \bibinfo{pages}{129--141}.
\newblock
\urldef\tempurl%
\url{https://doi.org/10.1016/j.imavis.2016.03.014}
\showDOI{\tempurl}


\bibitem[\protect\citeauthoryear{Salvucci and Goldberg}{Salvucci and
  Goldberg}{2000}]%
        {DBLP:conf/etra/SalvucciG00}
\bibfield{author}{\bibinfo{person}{Dario~D. Salvucci} {and}
  \bibinfo{person}{Joseph~H. Goldberg}.} \bibinfo{year}{2000}\natexlab{}.
\newblock \showarticletitle{Identifying fixations and saccades in eye-tracking
  protocols}. In \bibinfo{booktitle}{\emph{Proceedings of the Eye Tracking
  Research {\&} Application Symposium, {ETRA} 2000, Palm Beach Gardens,
  Florida, USA, November 6-8, 2000}}. \bibinfo{pages}{71--78}.
\newblock
\urldef\tempurl%
\url{https://doi.org/10.1145/355017.355028}
\showDOI{\tempurl}


\bibitem[\protect\citeauthoryear{Sargezeh, Tavakoli, and Daliri}{Sargezeh
  et~al\mbox{.}}{2019}]%
        {sargezeh2019gender}
\bibfield{author}{\bibinfo{person}{Bahman~Abdi Sargezeh},
  \bibinfo{person}{Niloofar Tavakoli}, {and} \bibinfo{person}{Mohammad~Reza
  Daliri}.} \bibinfo{year}{2019}\natexlab{}.
\newblock \showarticletitle{Gender-based eye movement differences in passive
  indoor picture viewing: An eye-tracking study}.
\newblock \bibinfo{journal}{\emph{Physiology \& behavior}}
  \bibinfo{volume}{206} (\bibinfo{year}{2019}), \bibinfo{pages}{43--50}.
\newblock


\bibitem[\protect\citeauthoryear{Sargolzaei, Abdelghani, Yen, and
  Sargolzaei}{Sargolzaei et~al\mbox{.}}{2016}]%
        {sargolzaei2016sensorimotor}
\bibfield{author}{\bibinfo{person}{Arman Sargolzaei}, \bibinfo{person}{Mohamed
  Abdelghani}, \bibinfo{person}{Kang~K. Yen}, {and} \bibinfo{person}{Saman
  Sargolzaei}.} \bibinfo{year}{2016}\natexlab{}.
\newblock \showarticletitle{Sensorimotor control: computing the immediate
  future from the delayed present}.
\newblock \bibinfo{journal}{\emph{BMC bioinformatics}} \bibinfo{volume}{17},
  \bibinfo{number}{7} (\bibinfo{year}{2016}), \bibinfo{pages}{501--509}.
\newblock


\bibitem[\protect\citeauthoryear{Savitzky and Golay}{Savitzky and
  Golay}{1964}]%
        {Savitzky1964}
\bibfield{author}{\bibinfo{person}{Abraham Savitzky} {and}
  \bibinfo{person}{Marcel J.~E. Golay}.} \bibinfo{year}{1964}\natexlab{}.
\newblock \showarticletitle{Smoothing and Differentiation of Data by Simplified
  Least Squares Procedures}.
\newblock \bibinfo{journal}{\emph{Analytical Chemistry}} \bibinfo{volume}{36},
  \bibinfo{number}{8} (\bibinfo{year}{1964}), \bibinfo{pages}{1627--1639}.
\newblock
\urldef\tempurl%
\url{https://doi.org/10.1021/ac60214a047}
\showDOI{\tempurl}
\showeprint{https://doi.org/10.1021/ac60214a047}


\bibitem[\protect\citeauthoryear{Schafer}{Schafer}{2011}]%
        {DBLP:journals/spm/Schafer11}
\bibfield{author}{\bibinfo{person}{Ronald~W. Schafer}.}
  \bibinfo{year}{2011}\natexlab{}.
\newblock \showarticletitle{What Is a Savitzky-Golay Filter? [Lecture Notes]}.
\newblock \bibinfo{journal}{\emph{{IEEE} Signal Process. Mag.}}
  \bibinfo{volume}{28}, \bibinfo{number}{4} (\bibinfo{year}{2011}),
  \bibinfo{pages}{111--117}.
\newblock
\urldef\tempurl%
\url{https://doi.org/10.1109/MSP.2011.941097}
\showDOI{\tempurl}


\bibitem[\protect\citeauthoryear{Schleicher, Galley, Briest, and
  Galley}{Schleicher et~al\mbox{.}}{2008}]%
        {schleicher2008blinks}
\bibfield{author}{\bibinfo{person}{Robert Schleicher}, \bibinfo{person}{Niels
  Galley}, \bibinfo{person}{Susanne Briest}, {and} \bibinfo{person}{Lars
  Galley}.} \bibinfo{year}{2008}\natexlab{}.
\newblock \showarticletitle{Blinks and saccades as indicators of fatigue in
  sleepiness warnings: looking tired?}
\newblock \bibinfo{journal}{\emph{Ergonomics}} \bibinfo{volume}{51},
  \bibinfo{number}{7} (\bibinfo{year}{2008}), \bibinfo{pages}{982--1010}.
\newblock


\bibitem[\protect\citeauthoryear{Schr{\"{o}}der, Zaidawi, Prinzler, Maneth, and
  Zachmann}{Schr{\"{o}}der et~al\mbox{.}}{2020}]%
        {DBLP:conf/chi/SchroderZPMZ20}
\bibfield{author}{\bibinfo{person}{Christoph Schr{\"{o}}der},
  \bibinfo{person}{Sahar Mahdie Klim~Al Zaidawi}, \bibinfo{person}{Martin H.~U.
  Prinzler}, \bibinfo{person}{Sebastian Maneth}, {and} \bibinfo{person}{Gabriel
  Zachmann}.} \bibinfo{year}{2020}\natexlab{}.
\newblock \showarticletitle{Robustness of Eye Movement Biometrics Against
  Varying Stimuli and Varying Trajectory Length}. In
  \bibinfo{booktitle}{\emph{{CHI} '20: {CHI} Conference on Human Factors in
  Computing Systems, Honolulu, HI, USA, April 25-30, 2020}}.
  \bibinfo{publisher}{{ACM}}, \bibinfo{pages}{1--7}.
\newblock
\urldef\tempurl%
\url{https://doi.org/10.1145/3313831.3376534}
\showDOI{\tempurl}


\bibitem[\protect\citeauthoryear{Sen and Megaw}{Sen and Megaw}{1984}]%
        {sen1984effects}
\bibfield{author}{\bibinfo{person}{Tayyar Sen} {and} \bibinfo{person}{Ted
  Megaw}.} \bibinfo{year}{1984}\natexlab{}.
\newblock \showarticletitle{The effects of task variables and prolonged
  performance on saccadic eye movement parameters}.
\newblock In \bibinfo{booktitle}{\emph{Advances in Psychology}}.
  Vol.~\bibinfo{volume}{22}. \bibinfo{publisher}{Elsevier},
  \bibinfo{pages}{103--111}.
\newblock


\bibitem[\protect\citeauthoryear{Wang, Woods, Costela, and Luo}{Wang
  et~al\mbox{.}}{2017}]%
        {wang2017dynamic}
\bibfield{author}{\bibinfo{person}{Shuhang Wang}, \bibinfo{person}{Russell~L.
  Woods}, \bibinfo{person}{Francisco~M. Costela}, {and} \bibinfo{person}{Gang
  Luo}.} \bibinfo{year}{2017}\natexlab{}.
\newblock \showarticletitle{Dynamic gaze-position prediction of saccadic eye
  movements using a Taylor series}.
\newblock \bibinfo{journal}{\emph{Journal of vision}} \bibinfo{volume}{17},
  \bibinfo{number}{14} (\bibinfo{year}{2017}), \bibinfo{pages}{1--11}.
\newblock
Issue 3.


\bibitem[\protect\citeauthoryear{Wang, Toor, Gautam, and Henson}{Wang
  et~al\mbox{.}}{2011}]%
        {wang2011blink}
\bibfield{author}{\bibinfo{person}{Yanfang Wang}, \bibinfo{person}{Sonia~S.
  Toor}, \bibinfo{person}{Ramesh Gautam}, {and} \bibinfo{person}{David~B.
  Henson}.} \bibinfo{year}{2011}\natexlab{}.
\newblock \showarticletitle{Blink frequency and duration during perimetry and
  their relationship to test--retest threshold variability}.
\newblock \bibinfo{journal}{\emph{Investigative ophthalmology \& visual
  science}} \bibinfo{volume}{52}, \bibinfo{number}{7} (\bibinfo{year}{2011}),
  \bibinfo{pages}{4546--4550}.
\newblock


\bibitem[\protect\citeauthoryear{Zaidawi, Prinzler, Schr{\"{o}}der, Zachmann,
  and Maneth}{Zaidawi et~al\mbox{.}}{2020}]%
        {DBLP:conf/icmi/ZaidawiPSZM20}
\bibfield{author}{\bibinfo{person}{Sahar Mahdie Klim~Al Zaidawi},
  \bibinfo{person}{Martin H.~U. Prinzler}, \bibinfo{person}{Christoph
  Schr{\"{o}}der}, \bibinfo{person}{Gabriel Zachmann}, {and}
  \bibinfo{person}{Sebastian Maneth}.} \bibinfo{year}{2020}\natexlab{}.
\newblock \showarticletitle{Gender Classification of Prepubescent Children via
  Eye Movements with Reading Stimuli}. In \bibinfo{booktitle}{\emph{Companion
  Publication of the 2020 International Conference on Multimodal Interaction,
  {ICMI} Companion 2020, Virtual Event, The Netherlands, October, 2020}}.
  \bibinfo{publisher}{{ACM}}, \bibinfo{pages}{1--6}.
\newblock
\urldef\tempurl%
\url{https://doi.org/10.1145/3395035.3425261}
\showDOI{\tempurl}


\end{thebibliography}
